\documentclass[journal]{IEEEtran}
\ifCLASSINFOpdf
\else
\fi
\usepackage{bm}
\usepackage{comment}
\usepackage{amsmath}
\usepackage{amssymb}
\usepackage{textcomp}
\usepackage{amsthm}
\usepackage{siunitx}
\usepackage{wrapfig}
\usepackage{todonotes}
\usepackage{enumitem}
\usepackage{siunitx}
\usepackage{dsfont}
\usepackage{xcolor}
\usepackage{tablefootnote}
\usepackage[para,online,flushleft]{threeparttable}
\usepackage{multirow}
\usepackage{mathtools}
\usepackage{booktabs}

\usepackage{algorithm}
\usepackage{algorithmicx}
\usepackage[noend]{algpseudocode}
\usepackage{changepage}
\usepackage{bbm}

\usepackage{subcaption} 
\usepackage{dblfloatfix}
\usepackage{balance}
\usepackage{diagbox}

\usepackage[comma, square, compress, numbers]{natbib}

\newcommand{\PreserveBackslash}[1]{\let\temp=\\#1\let\\=\temp}
\newcolumntype{C}[1]{>{\PreserveBackslash\centering}p{#1}}
\newcolumntype{R}[1]{>{\PreserveBackslash\raggedleft}p{#1}}
\newcolumntype{L}[1]{>{\PreserveBackslash\raggedright}p{#1}}

\begin{document}

%

\title{Enhanced Rolling Horizon Evolution Algorithm with Opponent Model Learning: 
Results for the Fighting Game AI Competition}

%
%
%

\author{Zhentao~Tang,~\IEEEmembership{Student~Member,~IEEE,}
        Yuanheng~Zhu,~\IEEEmembership{Member,~IEEE,}
        Dongbin~Zhao,~\IEEEmembership{Fellow,~IEEE,}
        and~Simon~M.~Lucas,~\IEEEmembership{Senior~Member,~IEEE}
\thanks{This work was supported in part by the National Key Research and Development Program of China under Grants 2018AAA0101005 and 2018AAA0102404.
	
	Z. Tang, Y. Zhu, D. Zhao are with the State Key Laboratory of Management and
	Control for Complex Systems, Institute of Automation, Chinese Academy
	of Sciences, Beijing 100190, China, and also with the School of Artificial Intelligence, University of Chinese Academy of Sciences, Beijing
	100049, China (e-mail: tangzhentao2016@ia.ac.cn; yuanheng.zhu@ia.ac.cn;
	dongbin.zhao@ia.ac.cn).
    
    S. M. Lucas is with the Department
    of Electronic Engineering and Computer Engineering (EECS), Queen Mary University of London, London E1 4NS, U.K. (e-mail:  simon.lucas@qmul.ac.uk).
}
}

%
%

\markboth{IEEE Transactions on games, ~2020}%
{Shell \MakeLowercase{\textit{et al.}}: Bare Demo of IEEEtran.cls for IEEE Journals}


%



\maketitle

\begin{abstract}

The Fighting Game AI Competition (FTGAIC) provides a challenging
benchmark for 2-player video game AI.  The challenge arises from the large action space,
diverse styles of characters and abilities, and the real-time nature of the game.
In this paper, we propose a novel algorithm that combines Rolling Horizon Evolution Algorithm (RHEA) with  opponent model learning.  The approach is readily applicable to any 2-player video game.
In contrast to conventional RHEA, an opponent  model is proposed and is optimized by supervised learning with cross-entropy and reinforcement learning with policy gradient and Q-learning respectively, based on history observations from opponent.  The model is learned during the live gameplay.
With the learned opponent model, the extended RHEA is able to make more realistic plans
based on what the opponent is likely to do.  This tends to lead to better results.
We compared our approach directly with the bots from the FTGAIC 2018 competition,
and found our method to significantly outperform all of them, for all three character.
Furthermore, our proposed bot with the policy-gradient-based opponent model is the 
only one without using Monte-Carlo Tree Search (MCTS) among top five bots in the 2019
competition in which it achieved second place, while using much less domain knowledge
than the winner.


\end{abstract}

\begin{IEEEkeywords}
Rolling horizon evolution, opponent model, reinforcement learning, supervised learning, fighting game.
\end{IEEEkeywords}

%
\IEEEpeerreviewmaketitle

\section{Introduction}
%
%
%
%
\IEEEPARstart{V}{ideo} games are able to model a range of real-world environments without much burden and unpredictable disruption, making them ideal testbeds for algorithms that might be slow or dangerous when performed in reality.  
Two-player zero-sum games have gained more and more attention from the game Artificial Intelligence (AI) community. Many related algorithms and frameworks have been proposed to address two-player game problems. Monte-Carlo Tree Search (MCTS), is one of the most famous algorithms and has been widely used in turn-based games, such as Go \cite{alphago,alphagozero}, Chess \cite{chess}, and Poker  \cite{poker}. MCTS-based approaches have reached or even surpassed top human players in recent years in many turn-based games, especially when combined with Deep Reinforcement Learning algorithms. 
Nevertheless, MCTS-based algorithms require a large number of iterations to select an effective planning path, and 
also have some computational overheads which has limited their application to real-time video games where
decisions are needed within a few tens of milliseconds.

In recent years, another algorithm named Rolling Horizon Evolution Algorithm (RHEA)  has successfully solved many real-time control tasks \cite{rhea-continue-control} and video games \cite{gvgai-2014, gvgai}. RHEA shortens the gap to MCTS and even outperforms in some games \cite{rhea-enhance}. In contrast to MCTS, RHEA uses a rolling horizon evolution technique to
search for optimal action sequences.  Compared to MCTS, RHEA has a lower computation overhead and
a better memory of its preferred course of action, which can make it better suited to
real-time video games.

\begin{figure}[htbp]
	\centering
	\includegraphics[width=6.5cm, height=4cm]{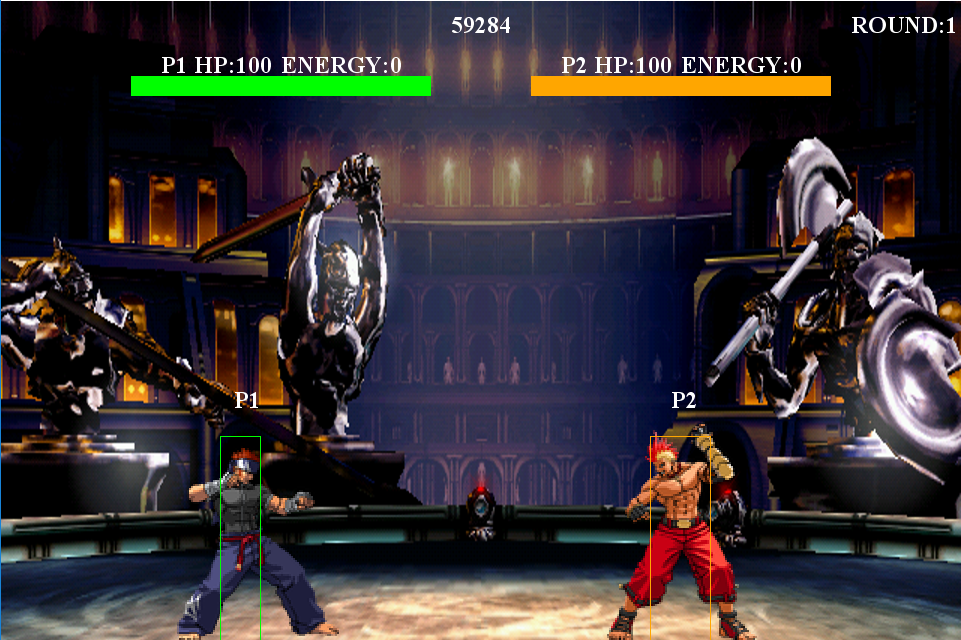}
	\caption{Screenshot of FightingICE.}
	\label{pic:fightingice}
\end{figure}

The FightingICE game platform is shown in Fig. \ref{pic:fightingice}, and was developed by the ICE Lab of Ritsumeiken University.  This has been used as the platform for the Fighting Game AI Competition (FTGAIC) series since 2015, with the platform undergoing a number of enhancements over the years to provide a faster and more robust forward model, and to overcome various exploits and provide a greater challenge.  

Each player chooses a character and takes actions, such as walk, jump, crouch, punch, kick, and guard to fight. 
The goal is to beat down the opponent and avoid being hit. Real-time demands (make a quick response in a short moment), incomplete information (simultaneous moves without the knowledge of opponent instant response from current observations), complex and changeable state-action space (each character has its own action space) are the main challenges of fighting game. In this work, FightingICE \cite{fightingICE} is chosen as our test platform, which particularly provides a simulator as the forward model for evolution planning. This simulator can be used as the model planner and provides a way to apply search-based algorithms in the game. 
But at each frame, there can be at most 56 actions for each character, and the search is required to be completed within 16.67ms.

In this paper, we first apply Rolling Horizon Evolution Algorithm to design a fighting AI. It optimizes the candidate action sequences through the evolutionary process via finite  iterations. Experiment results show that RHEA matches MCTS performance and even outperforms some MCTS-based bots. However, RHEA neglects opponent action selection, limiting the competitiveness of the optimized results. To deal with that, a variety of opponent models are introduced in RHEA, here we call it RHEAOM, to represent opponent action selection. A neural network is designed to infer the opponent next action according to its current state. The network is optimized based on new observations after each round. According to experimental results, the RHEAOM bot shows promising adaptability.  The learned model is adapted after each round of the game, so provides no advantage in the first round, but usually leads to a significant improvement in subsequent rounds.


The rest of this paper is organized as follows. Section \ref{related_work} briefly reviews related work of Fighting AI games. Section \ref{methods} presents the main algorithm and model proposed in this paper. Section \ref{experiments} describes the experimental setup and gives the results. Finally, Section \ref{conclusion} draws our conclusion and discusses future work.

\section{Related Work} \label{related_work} 


Scripted-based methods are widely used in game AI design and completely depend on human design and expert experience \cite{ai-games-book, ai-for-games}. For FightingICE,  \cite{ucb-script} adopts the UCB algorithm to select the rule-based controllers at a given time. \cite{script_agent} designs a real-time dynamic scripting AI, and it won the 2015 competition. 
Nevertheless, the scripted-based agent is constrained by limited cases and is easily exploited by opponents. 

MCTS-based agents have dominated FTGAIC since 2016. FightingICE platform provides a  built-in simulator and makes MCTS  applicable in real-time fighting game \cite{mcts-fight}. To take account of opponent behavior, \cite{opponet-fight} incorporates the manual action table for the opponent strategy modeling. But this method cannot defeat the 2016 FTGAIC winner, because it is under constraint by a small action table and fails in considering those more complicated cases.

Deep reinforcement learning (DRL) has demonstrated impressive results in many real-time video games \cite{review-drl, recent-drl, shao2019survey, starcraft, vizdoom, tang2018reinforcement}, such as Atari, Go, StarCraft, Dota2, and VizDoom. 
\cite{yoon2017deep} uses deep Q learning to show its potential for the two-player real-time fighting game. \cite{hra-fight} applies Hybrid Reward Architecture (HRA) based DRL into a fighting game AI. HRA decomposes the reward function into multiple components and learns the value functions separately. Though HRA-based DRL has shown promising performance, it is still defeated by MCTS-based AI GigaThunder, who is the champion of 2017 FTGAIC.

Opponent model approaches are mainly categorized as implicit and explicit opponent modeling. The implicit way is to maximize the  agent's own expected reward without having to  estimate the opponent's behavior \cite{he2016opponent}. While the explicit way  directly predicts the strategy of the opponent \cite{ganzfried2011game}. Compared with the implicit way, the explicit way is more efficient in training and more explainable in inference, and it is easier to combine with other existing algorithms.     For these reasons we use an explicit opponent model in this work.


\section{Main Algorithm and Model} \label{methods}
In this section, we propose a new algorithm that combines the rolling horizon evolution algorithm with an opponent learning model to design our real-time fighting game agent.

\subsection{Rolling Horizon Evolution Algorithm}

\begin{figure}[htbp]
	\centering
	\includegraphics[width=8cm, height=4cm]{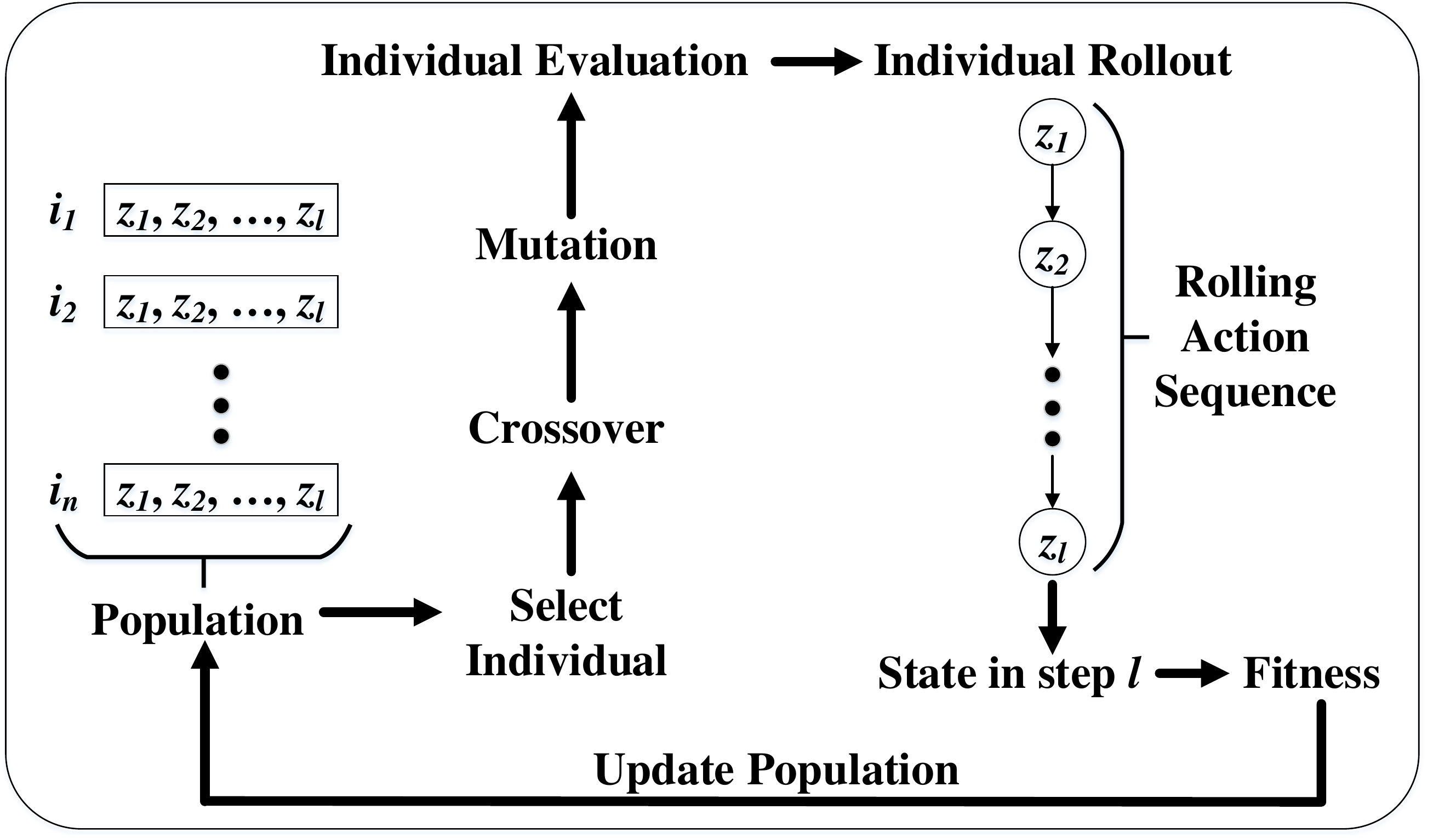}
	\caption{Flow diagram of RHEA.}
	\label{pic:rhea}
\end{figure}

RHEA is an optimization process that evolves action sequences through  forward model. After the optimization process, RHEA selects the first action of the sequence with the best fitness to perform in the task \cite{rhea-continue-control, rhea-process}. The flowchart of RHEA is shown in Fig. \ref{pic:rhea}. 

The population consists of multiple individuals that represent different action sequences. Each action in the sequence is viewed as a gene. A certain number of individuals are selected according to the fitnesses. Then, these individuals are assigned to crossover and there is a probability of mutation to generate more potential or more powerful offspring. Afterwards, individuals are rolled out as the action sequences and the action sequences are inputted orderly into forward model to infer the future states. The future states are evaluated by the score function to obtain new fitness. The above optimization process is repeated until the time budget is consumed.

In our implementation of RHEA, the action sequence in each individual is initialized randomly. After initialization, a group of individuals are created with the same length, and each individual is evaluated by the fitness function as

\begin{equation}
f_{fit}(s_t, \vec{z}^{l}, \vec{o}^{l}) = (1 - \lambda)f_{sco}(f_{FM}(s_t, \vec{z}^{l}, \vec{o}^{l})) + \lambda f_{div}(\vec{z}^{l}),
\end{equation}
where
\begin{align}
\hspace{0.35cm}  f&_{sco}(s_t)=\begin{cases}
-1, \emph{ \hspace{0.5cm}  if end with loss}\\
1,  \emph{ \hspace{0.8cm}  if end with win}\\
\frac{1}{\max{(hp)}}(hp_{self}(s_t) - hp_{opp}(s_t)), \emph{ otherwise} \\ 
\end{cases} \label{eq:score} \\
  f&_{div}(\vec{z}^{l}) = 1 - \frac{1}{nl}\sum\nolimits_{j=1}^{l}{f_{count}(\vec{z}^{l}(j))}, \label{eq:diversity}  \\
  f&_{FM}(s_t, \vec{z}^{l}, \vec{o}^l) = s_{t+l}, \label{eq:fm}
\end{align}
here $\lambda$ is the weight for diversity and score. {$\vec{z}^{l}$} represents the action sequence of an individual and {$\vec{o}^{l}$} is the opponent action sequence both with length {$l$}. {$s_t$} is the state at timestep {$t$}. {$hp$} is the hit point of corresponding player.  {$f_{fit}$} is the fitness function, weighted averaged on score function {$f_{sco}$} and diversity function {$f_{div}$}. The score function $f_{sco}$ is used to evaluate the value of $s_t$, and the diversity function $f_{div}$ is used to avoid falling into the local optimal solution. {$n$} is the population size. {$\vec{z}^l(j)$} is the $j^{th}$ action in  sequence. {$f_{count}$} is the count function of the occurrence of one gene in current population. {$f_{FM}$} is the forward model that determines the future state according to the current state and the action sequences. 

Suppose we have a total of $n$ individuals in current generation. Top $k$ highest scored individuals are picked as elites and are preserved in the next generation. The remained $n-k$ individuals are evolved with elites by crossover with the uniform cross operator, and  parents are randomly drawn from elites and remained individuals respectively.
Afterwards, one gene is selected randomly from the individual, and it is mutated into another valid gene through a uniform distribution. Finally, these latest individuals are reevaluated by the fitness function. If there is still a time budget, select top $n$ sorted individuals for next generation and repeat the above evolution again. Otherwise, command the first action from the highest sorted individual. The whole process of rolling horizon evolution algorithm with the opponent model for the fighting game is given in Algorithm \ref{alg:rheaom}.

\renewcommand{\algorithmicensure}{\textbf{Output:}}

\begin{algorithm}[h]
	\caption{Rolling Horizon Evolution Algorithm with Opponent Model (RHEAOM) for Fighting Game.}
	\label{alg:rheaom}
	\begin{algorithmic} [0]
		\Require{$\mathbb{A}$: candidate action set.\\}
		{$n, k\in N^{+}$: number of population and elites.\\}
		{$p_m\in(0,1)$: threshold of mutation probability.\\}
		{$\lambda\in(0,1)$: weight for score and diversity.\\}
		{$OM$: an opponent model to infer enemy action. }
		\Ensure{action for the fighting game.}
		\State{}
		\State{$\mathbb{Z}^{n \times l} \leftarrow$ randomly generates $n$ action sequences $\vec{z}^l$ with length $l$ from $\mathbb{A}$. }
		\State{$\vec{v}^{n} \leftarrow Evaluate(Forward(\mathbb{Z}^{n \times l}, OM), \mathbb{Z}^{n \times l})$}
		\State{$\mathbb{Z}^{n \times l} \leftarrow$ sort $\mathbb{Z}^{n \times l}$ according to $\vec{v}^{n}$ from high to low.}
		\While{$ time~budget \le remained~time $}
		\State{$\mathbb{Z}_{el}^{k \times l} \leftarrow$ top $k$ elites from $\mathbb{Z}^{n\times l}$.}
		\State{$\mathbb{Z}_{re}^{(n-k) \times l} \leftarrow$ rest $(n-k)$ individuals from $\mathbb{Z}^{n\times l}$.}
		\State{$\mathbb{Z}_{new}^{(n-k)\times l} \leftarrow$ create $(n-k)$ new individuals, by uniformly crossover individual $i_1$ and $i_2$, where $i_1$ from $\mathbb{Z}_{el}^{k \times l}$ and $i_2$ from $\mathbb{Z}_{re}^{(n-k)\times l}$ for ($n-k$) times}
		\If{mutation probability $<p_m$}
		\State{$\mathbb{Z}_{new}^{(n-k)\times l} \leftarrow$ mutate one gene from $\mathbb{Z}_{new}^{(n-k)\times l}$.} 
		\EndIf
		\State{$\mathbb{Z}^{n \times l} \leftarrow \mathbb{Z}_{el}^{k \times l} \cup \mathbb{Z}_{new}^{(n-k)\times l}$}
		\State{$\vec{v}^{n} \leftarrow Evaluate(Forward(\mathbb{Z}^{n \times l}, OM), \mathbb{Z}^{n \times l})$}
		\State{$\mathbb{Z}^{n \times l} \leftarrow$ sort $\mathbb{Z}^{n \times l}$ by $\vec{v}^{n}$ from high to low.}
		\EndWhile
		\State{$\vec{a}^{l} \leftarrow$ choose the highest sorted action sequence from  $\mathbb{Z}^{n\times l}$.  }
		\State{\Return {$\vec{a}^{l}(1)$}\Comment{The first action of sequence.}}
		
		\Statex
		\Function{Forward}{$\mathbb{Z}^{n\times l}, OM$}
		\For{$i \in \{1,\dots,n\}$}
		\State{$\vec{s}_{t}^{n}(i) \leftarrow curFrame$} \Comment{Initialize as current frame.}
		\For{$j \in \{1,\dots,l\}$}
		\State{$a_{o} \leftarrow OM(\vec{s}_{t}^{n}(i))$} \Comment{Infer the opponent action.}
		\State{$\vec{s}_{t}^{n}(i) \leftarrow f_{FM}(\vec{s}_{t}^{n}(i), \mathbb{Z}^{n\times l}(i,j), a_{o})$ \Comment{(\ref{eq:fm})}}
		\EndFor
		\EndFor
		\State{\Return{$\vec{s}_{t}^{n}$} \Comment{Each action sequence's final frame.}}
		\EndFunction
		
		\Statex
		\Function{Evaluate}{$\vec{s}_{t}^{n}, \mathbb{Z}^{n \times l}$}
		
		\State{$\vec{p}_{s}^{n} \leftarrow f_{sco}(\vec{s}_{t}^{n})$ \Comment{Score evaluation (\ref{eq:score}).}} 
		\State{$\vec{p}_{d}^{n} \leftarrow f_{div}(\mathbb{Z}^{n \times l})$ \Comment{Diversity evaluation (\ref{eq:diversity}).}}
		\State{$\vec{v}^{n} \leftarrow (1-\lambda) (\vec{p}_{s}^{n}) + \lambda (\vec{p}_{d}^{n})$} 
		\State{\Return{$\vec{v}^{n}$ \Comment{Fitness evaluation.}}}
		\EndFunction
		
	\end{algorithmic}
\end{algorithm}

In our fitness definition, a criterion of action occurrence frequency is included, and it reflects the gene diversity in population.  
According to (\ref{eq:diversity}), high diversity is preferred, it is mainly because high diversity helps explore more feasible solutions and avoids stuck in a local optimum.     
\begin{figure*}[htbp]
	\centering
	\includegraphics[height=10.5cm, width=17.5cm]{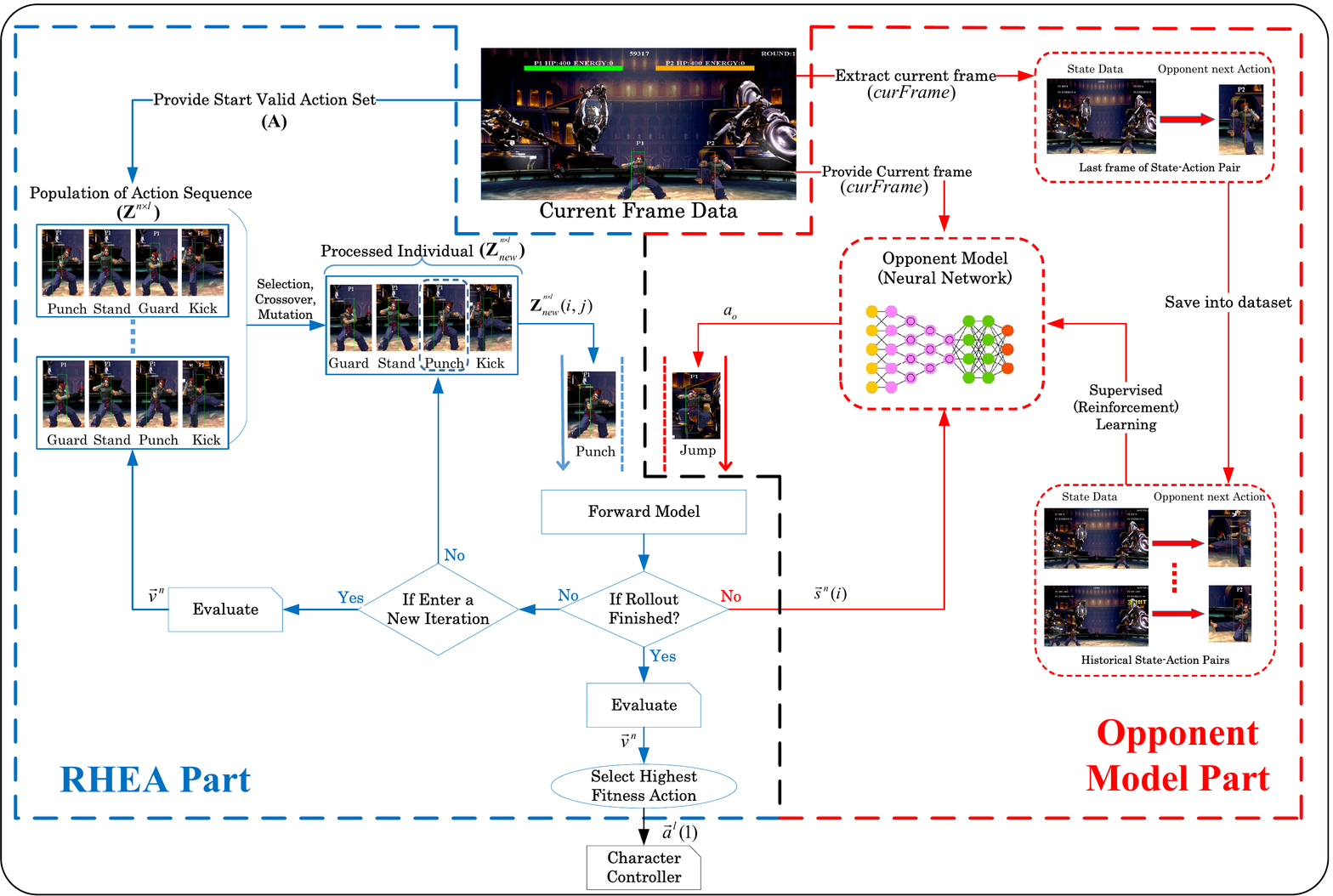}
	\caption{Flow diagram of RHEAOM for Fighting Game AI.}
	\label{pic:process}
\end{figure*}

\subsection{Opponent Learning Model}
Since RHEA merely considers future behavior of itself, it cannot directly infer which action will be taken by the opponent. Obviously, it will mislead the evaluation with only self-consideration. Despite \cite{opponet-fight} has incorporated an opponent model in MCTS by setting up an activity table as its opponent model, such a model does not lead to a significant improvement in  performance.  

To deal with that, we propose an opponent learning model for the real-time fighting game, and it is conducive to RHEA with more credible rolling horizon evolution. The opponent learning model uses a one-step look-ahead for the opponent behavior inference and learning. Inspired by the excellent fitting performance of neural networks, the neural-network-based model is constructed as the opponent model. 

\vspace{0.2cm}
\noindent \textbf{B.1. Supervised Learning based Opponent Model} 
\vspace{0.1cm}

Under the real-time restriction and memory limitation, a simple neural network model is suitable for this task. It has 18 numerical inputs, including both hit point, energy, coordinate of the arena in x-axis and y-axis, state of the character, and the relative distance between two characters in the directions of x and y respective. The detail of input features will be described in the Section \ref{experiments}. The output of the layer has a total of 56 nodes corresponding to all actions in FightingICE. The loss function is the standard cross-entropy as 
\begin{equation}
\mathcal{L}_{ce}=-\sum\nolimits_{j=1}^{l}{p_j \log (q_j)},
\end{equation}
where $p_j$ is the actual action label from the opponent history as one-hot vector form, and $q_j$ is the network output. In contrast to conventional off-line supervised learning for recognition tasks, this neural network based opponent model uses on-line training with the latest observations for adaption against different opponents. 

\vspace{0.2cm}
\noindent \textbf{ B.2. Reinforcement Learning based Opponent Model}
\vspace{0.1cm}

In addition to use  cross-entropy-based supervised learning for the  opponent model, we also adopt the reinforcement learning based on reinforcement learning: Q-learning \cite{Q-learning} and policy gradient \cite{policy-gradient}. In the traditional reinforcement learning setting, agents should sample the action to interact with the environment through some techniques, likes $\epsilon$-greedy, UCB, and  Thompson Sampling. But here the reinforcement learning is only for evolution in RHEA and the opponent has its own policy to interact with the fighting game. 

\subsubsection{\textbf{Q-Learning based Opponent Model}}
In order to accelerate the convergence of the learning process, the training goal is the $N$-step return: 
\begin{equation} \label{eq:n_step}
  G_{t}^{(N)}=\sum_{l=1}^{N}{\gamma^{l-1}r_{t+l} + \gamma^{N}f^{Q}(s_{t+N}, \theta^{Q})}. 
\end{equation}
The opponent model parameters $\theta^{Q}$ are updated by minibatch gradient descent to minimize the mean-square loss  
\begin{equation} \label{eq:q}
\mathcal{L}_t(\theta^{Q}) = ( G_{t}^{(n)} - f^Q(s_t, \theta^Q))^{2}.
\end{equation}

\subsubsection{\textbf{Policy-Gradient based Opponent Model}}
We also adopt another typical reinforcement learning update rule named policy gradient. This method directly optimizes an agent policy, which is parameterized by $\theta^{\pi}$, by performing gradient ascent on the estimation of the expected discount total reward $J=\mathbb{E}_{\pi}[R_{0}]$ from scratch. The gradient of the policy-gradient method is 
\begin{equation} \label{eq:pi} 
g=\mathbb{E}_{s_t, a_t}[\sum_{t=0}^{T}{R_{t} \nabla_{\theta^{\pi}} \log \pi(a_t|s_t)}],
\end{equation}
where the cumulative reward  $R_t=\sum_{l=t}^{T}{\gamma^{l-t}r_{l}}$. 
It is noteworthy that reward $r_t$ in (\ref{eq:n_step}) and  (\ref{eq:pi}) is the hit point difference between the two sides and is defined as
\begin{equation} \label{eq:r}
r_t = (hp_{t+1}^{opp} - hp_{t+1}^{self}) / \max{(hp)},
\end{equation}
where $\max{(hp)}$ is the initial hit point of player. 
Positive reward $r_t$ means the opponent has more hit point than self-own at the next state and vice versa.

With the aid of the opponent model, there is a fictitious player to generate opponent  behaviors. RHEAOM is able to make more resultful evaluation for self-action sequences. The flow diagram is presented in Fig.  \ref{pic:process}. The notations in Fig. \ref{pic:process} are kept the same as shown in Algorithm \ref{alg:rheaom}, except $Z_{new}^{n\times l}(i,j)$ denotes the $j^{th}$ action of the $i^{th}$ processed individual.

\section{Experiments} \label{experiments}

\subsection{Experiments Setup}
In this section, we introduce the fighting game AI platform (FightingICE) to which we apply RHEAOM, and describes details of state features of opponent model, network architecture, and training regimes.

\textbf{Platform \& Setup.} FightingICE provides a real-time fighting game platform and it is suitable for fighting game AI test. The player is required to choose one of 56 actions to perform in FightingICE within 16.67ms each frame. To simulate the reaction delay of human players, FightingICE has 15 frames delay which means the bot is accessible to only 15 earlier frames. Furthermore, there are three different characters in FightingICE, ZEN, GARNET (GAR) and LUD. These three characters have entirely different strong combinations of actions toward the same state. In summary, FightingICE gives the challenges on real-time planning and decision, simultaneous moves of two sides with time-delay observation, and generalization of various characters.        

Both two sides at the fighting test are forced to use the same character by the rule of FTGAIC for a fair play. The goal of each player is to defeat the opponent within a limited time, which is 60s in a round. 
In this game, we design the agent to choose from a set of discrete actions: {\textit{Stand}}, {\textit{Dash}}, {\textit{Crouch}}, {\textit{Jump[direction]}}, {\textit{Guard}}, {\textit{Walk[direction]}}, {\textit{Air Attack[type]}}, {\textit{Ground Attack[type]}}, and so forth. In the FightingICE, the action execution of  each player contains three stages: {\textit{startup}}, {\textit{active}} and {\textit{recover}}.  It indicates that the action has to spend a certain number of frames to be executed. Once the action is taken, then it cannot be interrupted unless the player is under attack by the opponent.  

Comparative evaluations are set up to verify the performance of RHEA agents with and without the opponent model by self comparison and versus 2018 FTGAIC bots. In consideration of the game balance, we  select the same character GAR, LUD, and ZEN for both sides. Each opponent model is initialized randomly and has been examined in 200 rounds with five times repeatedly. According to the rule of FTGAIC, initial hit point of each character is set to 400 and the initial energy is set to 0. Each round is ended if hit point of either player reaches 0 or 60 seconds elapse. The one who has higher hit point is the winner of the round. The relative hyper-parameters of RHEAOM are demonstrated in Table \ref{tab:hyper-table}.

\begin{table}[h]
	\begin{center}
		\begin{tabular}{ccl}
			\hline
			Parameters & value  & \multicolumn{1}{c}{description} \\
			\hline
			$\lambda$  &  0.5 & Weight for diversity and score. \\
			$lr$      &  1e-4 & Learning rate. \\
			$p_m$  &  0.85 & Probability of mutation. \\
			$n$  &  7   & The number of the total individuals.	\\
			$l$    &  4   & Length of the action sequence. \\
			$\#elite$	 &  1   & The number of the elite in the population.  \\
			
			\hline       
		\end{tabular}
	\end{center}
	\caption{Hyper-parameters Set for RHEAOM Agent}
	\label{tab:hyper-table}
\end{table}

In order to meet the real-time requirement, the hyper-parameters are relatively small except the probability of mutation, since this higher probability will encourage to explore a better solution. Besides, we have tried the technique called \textit{Shift Buffer}, but it does not make much difference since it may be suitable for long-horizon planning and the length of the action sequence here is relatively small.

\begin{table*}[h]
	\centering
	\begin{tabular}{c|C{0.6cm}C{0.6cm}C{0.6cm}|C{0.6cm}C{0.6cm}C{0.6cm}|C{0.6cm}C{0.6cm}C{0.6cm}|C{0.6cm}C{0.6cm}C{0.6cm}|C{0.6cm}C{0.6cm}C{0.6cm}}
		\toprule
		\diagbox{Opps}{Ours} & \multicolumn{3}{c|}{RHEA(\%)} & \multicolumn{3}{c|}{RHEAOM-R(\%)} & \multicolumn{3}{c|}{RHEAOM-SL(\%)} & \multicolumn{3}{c|}{RHEAOM-Q(\%)} & \multicolumn{3}{c}{RHEAOM-PG(\%)} \\
		\midrule
		Chars   & GAR   & LUD   & ZEN   & GAR   & LUD   & ZEN   & GAR   & LUD   & ZEN   & GAR   & LUD   & ZEN   & GAR   & LUD   & ZEN \\
		\cmidrule{1-16}    \multicolumn{1}{c|}{N} & -     & -     & \multicolumn{1}{c|}{-} & \textbf{58.2}(3)  & 59.7(3)  & \multicolumn{1}{c|}{56.3(3)} & 50.9(3)  & \textbf{82.2}(1)  & \multicolumn{1}{c|}{67.8(3) } & 34.2(4)  & 80.1(4)  & \multicolumn{1}{c|}{72.2(3) } & 51.5(4)  & 81.9(6)  & \textbf{79.3}(3)  \\
		\multicolumn{1}{c|}{R} & 41.8(3)  & 40.3(3)  & \multicolumn{1}{c|}{43.7(3)} & -     & -     & \multicolumn{1}{c|}{-} & 49.0(4)  & 61.4(3)  & \multicolumn{1}{c|}{61.5(2) } & 43.4(3)  & 64.8(3)  & \multicolumn{1}{c|}{65.5(4) } & \textbf{52.0}(3)  & \textbf{68.0}(2)  & \textbf{66.7}(5)  \\
		\multicolumn{1}{c|}{SL} & 49.1(3)  & 17.8(1)  & \multicolumn{1}{c|}{32.2(3) } & 51.0(4)  & 38.6(3)  & \multicolumn{1}{c|}{38.5(2) } & -     & -     & \multicolumn{1}{c|}{-} & 46.2(3)  & 47.4(3)  & \multicolumn{1}{c|}{51.2(3) } & \textbf{53.8}(4)  & \textbf{52.0}(3)  & \textbf{59.4}(3)  \\
		\multicolumn{1}{c|}{Q} & \textbf{65.8}(4)  & 20.0(4)  & \multicolumn{1}{c|}{27.9(3)} & 56.6(3)  & 35.2(3)  & \multicolumn{1}{c|}{34.5(4) } & 53.9(3)  & 52.6(3)  & \multicolumn{1}{c|}{48.8(3) } & -     & -     & \multicolumn{1}{c|}{-} & 54.1(5)  & \textbf{53.0}(4)  & \textbf{54.2}(4)  \\
		\multicolumn{1}{c|}{PG} & \textbf{48.5}(4)  & 18.1(6)  & \multicolumn{1}{c|}{20.7(3)} & 48.0(3)  & 32.0(2)  & \multicolumn{1}{c|}{33.3(5) } & 46.2(4)  & \textbf{48.1}(3)  & \multicolumn{1}{c|}{40.6(3) } & 45.9(5)  & 47.0(4)  & \multicolumn{1}{c|}{\textbf{45.9}(4) } & -     & -     & - \\
		\cmidrule{1-16}    Mean  & 51.3  & 24.1  & 31.1  & \textbf{53.5}  & 41.4  & 40.7  & 50.0  & 61.1  & 54.7  & 42.4  & 59.8  & 58.7  & 52.8  & \textbf{63.7}  & \textbf{64.9}  \\
		\bottomrule
	\end{tabular}%
	\caption{\textbf{Win rate of self comparison for models in first row.} Every test is repeated for five times to average the win rate. These are the five variant opponent models with RHEA, \texttt{N:None, R:Random, SL:Supervised Learning, Q:Q-Learning, PG:Policy Gradient}, and the last \texttt{Mean} row is the average win rates over the four opponents. The highest values are in bold. The values in parentheses denote the 95\% confidence interval, for example 58.2(3)=58.2$\pm$3. }
	
	\label{tab:self-comparison}%
\end{table*}%

\textbf{State Features of Opponent Model.} Opponent model is used to estimate which action will be most probably and effectively taken by the opponent.
There are a total of 18 input features for network and their details are as follows:
\begin{itemize}
	\item {\emph{Hit point (1-2)}}, hit points of p1 and p2. 
	\item {\emph{Energy (3-4)}}, energies of p1 and p2.
	\item {\emph{Coordinate x and y (5-8)}}, locations of p1 and p2. 
	\item {\emph{States (9-16)}}, one-hot encoding for character states (Stand, Crouch, Air, Down) of p1 and p2.
	\item {\emph{Distance (17-18)}}, relative distance between p1 and p2 in the directions of x-axis and y-axis.
\end{itemize}
Here p1 and p2 respectively represent player 1 and player 2. 
All these input features are normalized into [0, 1] by their maximum values, such as the maximum of hit point and energy, the width and height of the fighting stage for x-axis and y-axis, except the one-hot encoding for character state. 
It is noteworthy that these game states are provided by the game engine with a certain number of delayed frames, so there is certain bias if bot directly uses these states to make decision. In order to address this problem, we adopt the forward model to plan the next state with currently processed state.   

\textbf{Architecture \& Training.} Since RHEAOM optimizes the solution through evolution and iteration, the opponent model has to be as simple and concise as  possible for fast and multiple iterations. 
The opponent model only consists of a single input and output layer without any hidden layer. As mentioned above there are 18-bit units (State Feature) as input layer and 56-bit units (Discrete Action Set) as output layer. These two layers are fully connected with the linear activation function. In addition, discrete action distribution is generated from the output layer via the Softmax function. We adopt XAVIER \cite{xavier} for the network initialization and Adam \cite{Adam} as the network optimizer. In addition, other more complicated network architectures, such as multilayer perceptron  and long short term memory network, have been tested but their performance is worse than the simple architecture.  

In each round, opponent state-action pairs are first recorded in a dataset. At the end of the round, the opponent model is trained by the latest dataset. There are about five seconds for the preparation of  next round. Once a new round starts, the dataset is emptied. There are two reasons for postponing the training at the round end. First, it concedes the time budget for action decision by RHEA at each frame. Secondly, it helps improve the training stability and reliability compared to instant update after each frame.

\subsection{Self Comparison}
We perform comparative evaluations to validate the three key factors of RHEAOM. First, we test the effect of the opponent model in the RHEA framework by comparing it against the None-opponent model and the Random-opponent model. The None-opponent model means that we assume the opponent does not take any action but just stands on the ground, while the Random-opponent model means that the opponent's action is randomly sampled from the valid candidate action set. For instance, the opponent cannot take any ground action when it is in the air, so the candidate valid action set consists of just air actions, such as flying attack and flying guard.  
Second, we verify the effect of different training rules for the opponent model, including cross-entropy-based supervised learning, Q-learning-based, and policy-gradient-based reinforcement learning.  
Third, we observe the winning convergence efficiency of whether using an opponent learning model or not for three characters. We also verify whether the opponent model can aid to accelerate the reach to equilibrium, that is the equal performance for both sides.

Here we set up five variant versions of RHEA to test their performance,
\begin{itemize}
	\item \textbf{\emph{RHEA}}, vanilla RHEA without opponent model. 
	\item \textbf{\emph{RHEAOM-R}}, RHEA combines with random opponent model.
	\item \textbf{\emph{RHEAOM-SL}}, RHEA combines with supervised-learning-based opponent model. 
	\item \textbf{\emph{RHEAOM-Q}}, RHEA combines with Q-learning-based opponent model.
	\item \textbf{\emph{RHEAOM-PG}}, RHEA combines with policy-gradient-based opponent model.
\end{itemize}
All these RHEA-based algorithms are tested against other variants except themselves. As shown in Table  \ref{tab:self-comparison}, RHEAOM-PG has superior performance for three characters when fighting against other variants. Though RHEAOM-PG does not achieve the highest mean win rate as GAR, it still exceeds other models over 50 percent for three characters, and it defeats RHEAOM-R which has the highest mean win rate as GAR. 

\begin{figure}[H]
	\centering
	\includegraphics[width=8cm, height=4.5cm]{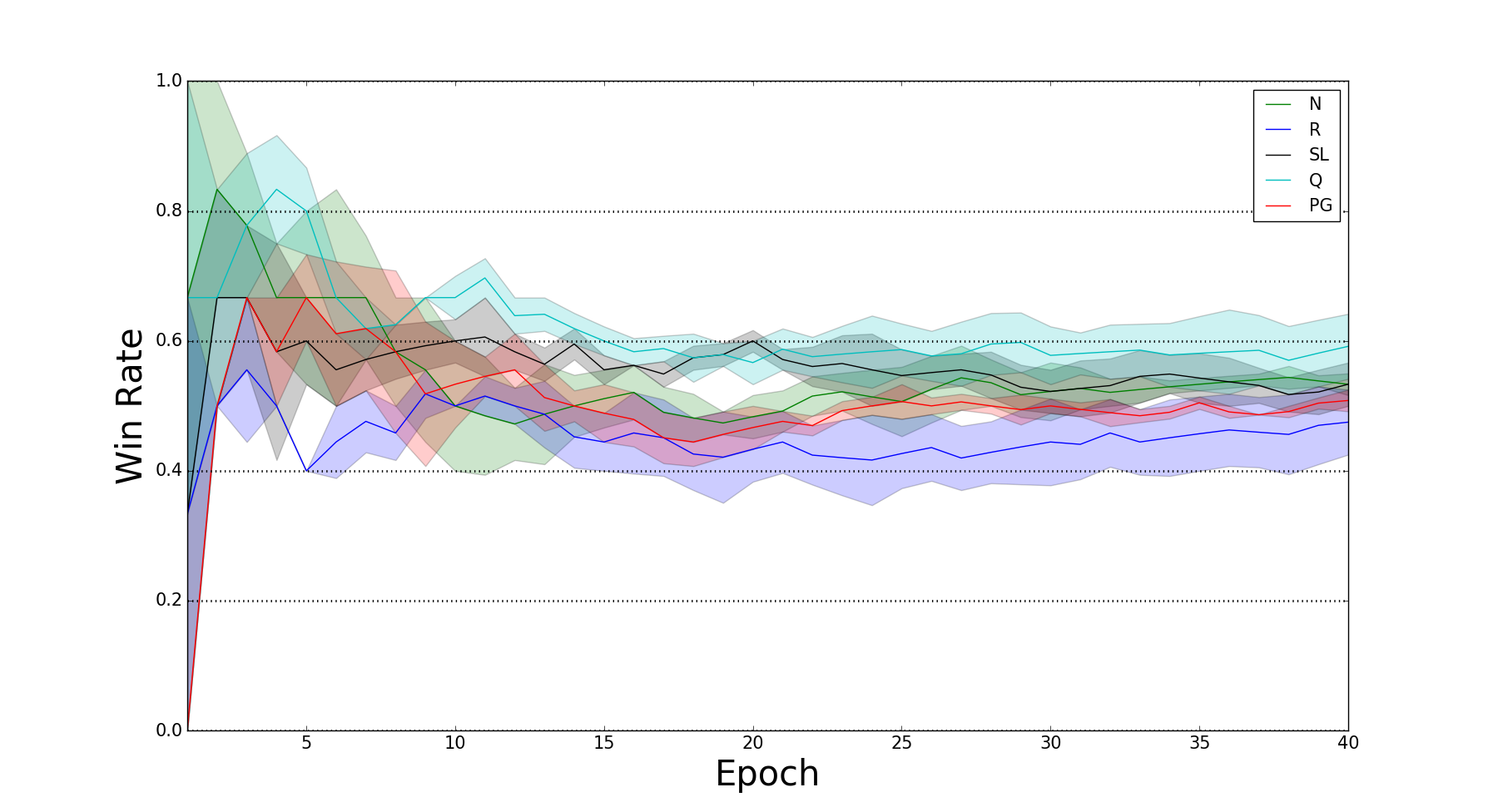}
	\caption{The curves of the average win rate means RHEA-based bots fight against itself with the increasing of iterations.}
	\label{pic:plot}
\end{figure}

In order to test the convergence efficiency of self-play, we set up RHEA with and without opponent model to fight against itself for all three characters. According to the experiment results in Table \ref{tab:self-comparison}, RHEAOM-PG performs better than other RHEA-based approaches. In order to observe the convergence curve, we select RHEA, RHEAOM-R, RHEAOM-SL, RHEAOM-Q, and RHEAOM-PG for comparisons. As presented in Fig. \ref{pic:plot}, the win rate converges through the iterative process. RHEA and RHEAOM-R show more obvious vibration in terms of win rate at the early and middle phases than RHEAOM-PG. RHEAOM-Q presents obvious vibration at the final phase. This experiment also shows that RHEAOM-SL and RHEAOM-PG converges faster to their equilibriums than other variants.

\begin{table*}[h]
	\centering
	\begin{tabular}{c|C{0.6cm}C{0.6cm}C{0.6cm}|C{0.6cm}C{0.6cm}C{0.6cm}|C{0.6cm}C{0.6cm}C{0.6cm}|C{0.6cm}C{0.6cm}C{0.6cm}|C{0.6cm}C{0.6cm}C{0.6cm}}
		\toprule
		\diagbox{Opps}{Ours} & \multicolumn{3}{c|}{RHEA(\%)} & \multicolumn{3}{c|}{RHEAOM-R(\%)} & \multicolumn{3}{c|}{RHEAOM-SL(\%)} & \multicolumn{3}{c|}{RHEAOM-Q(\%)} & \multicolumn{3}{c}{RHEAOM-PG(\%)} \\
		\cmidrule{1-16}    Chars & GAR   & LUD   & ZEN   & GAR   & LUD   & ZEN   & GAR   & LUD   & ZEN   & GAR   & LUD   & ZEN   & GAR   & LUD   & ZEN \\
		\midrule
		\multicolumn{1}{c|}{Th} & 82.3(6)  & 62.3(4)  & \multicolumn{1}{c|}{70.5(7) } & 78.3(6)  & 62.2(3)  & \multicolumn{1}{c|}{50.5(5) } & 84.0(2)  & 83.6(2)  & \multicolumn{1}{c|}{69.8(2) } & 84.5(5)  & 88.5(2)  & \multicolumn{1}{c|}{69.0(3) } & \textbf{89.8}(4)  & \textbf{89.3}(1)  & \textbf{76.4}(3)  \\
		\multicolumn{1}{c|}{KT} & 66.0(7)  & 90.3(3)  & \multicolumn{1}{c|}{ \textbf{88.9}(4)} & 74.0(3)  & 86.9(2)  & \multicolumn{1}{c|}{11.3(1) } & \textbf{78.6}(2)  & \textbf{95.8}(1)  & \multicolumn{1}{c|}{77.1(2) } & 66.1(3)  & 92.2(1)  & \multicolumn{1}{c|}{72.8(6) } & 78.2(4)  & 92.5(1)  & 72.1(7)  \\
		\multicolumn{1}{c|}{Jay} & 98.5(1)  & 62.3(4)  & \multicolumn{1}{c|}{63.7(1) } & \textbf{100.0}  & 81.5(2)  & \multicolumn{1}{c|}{64.3(4) } & 96.6(2)  & 91.0(3)  & \multicolumn{1}{c|}{87.8(8) } & 98.0(1)  & \textbf{94.7}(2)  & \multicolumn{1}{c|}{92.0(2) } & 98.0(1)  & 94.6(1)  & \textbf{97.0}(1)  \\
		\multicolumn{1}{c|}{Mo} & 78.5(2)  & 64.0(1)  & \multicolumn{1}{c|}{70.0(6) } & 77.1(2)  & 81.6(2)  & \multicolumn{1}{c|}{39.1(6) } & 80.8(6)  & \textbf{90.9}(3)  & \multicolumn{1}{c|}{88.5(1) } & \textbf{83.6}(2)  & 87.2(3)  & \multicolumn{1}{c|}{82.8(2) } & 82.1(3)  & 89.0(3)  & \textbf{93.8}(4)  \\
		\multicolumn{1}{c|}{Ut} & 97.3(1)  & 69.1(2)  & \multicolumn{1}{c|}{73.9(6) } & 96.8(1)  & 87.3(3)  & \multicolumn{1}{c|}{87.9(1) } & 98.1(1)  & 94.6(1)  & \multicolumn{1}{c|}{96.9(2) } & 99.7(1)  & 98.6(2)  & \multicolumn{1}{c|}{96.9(2) } & \textbf{100.0}  & \textbf{100.0}  & \textbf{97.1}(1)  \\
		\midrule
		Mean  & 84.5  & 69.6  & 73.4  & 85.2  & 79.9  & 50.6  & 87.6  & 91.2  & 84.0  & 86.4  & 92.3  & 82.7  & \textbf{89.6}  & \textbf{93.1}  & \textbf{87.3}  \\
		\bottomrule
	\end{tabular}%
	\caption{\textbf{Win rate against 2018 FTGAIC Bots.} Every test is repeated for  five times to average the win rate. These are the five java-based bots from 2018 FTGAIC, \texttt{Th:Thunder, KT:KotlinTestAgent, Jay:JayBotGM, Mo:MogakuMono, Ut:UtalFighter}. The last \texttt{Mean} row denotes the average win rates over the five competition bots. The highest values are in bold. The values in parentheses denote the 95\% confidence interval. }
	\label{tab:vs-2018}%
\end{table*}%

\subsection{Versus 2018 FTGAIC Bots}
To measure the performance of our proposed frameworks in FTGAIC, we choose five Java-based bots from the 2018 FTGAIC as opponents. Since FightingICE is equipped with a simulator that can be considered as the forward model, most bots are designed based on  MCTS. The five bots considered here are listed below according to their ranks in 2018 FTGAIC:
\begin{itemize}
	\item \textbf{\emph{Thunder}}, $1^{st}$, based on MCTS with different heuristic settings towards different characters. 
	\item \textbf{\emph{KotlinTestAgent}}, $2^{nd}$, utilizes a hybrid solution based on MCTS selection optimization and smart corner case strategy.
	\item \textbf{\emph{JayBotGM}} \cite{jaybot}, $3^{rd}$, makes use of the combination of genetic algorithm and MCTS. 
	\item \textbf{\emph{MogakuMono}}, $4^{th}$, conducts a hierarchical reinforcement learning framework into the fighting agent.
	\item \textbf{\emph{UtalFighter}}, $8^{th}$, adopts a simple finite state machine to make a decision.
\end{itemize}

Due to UtalFighter is a script-based agent, it can be directly used to represent the learning curve of our proposed opponent learning models.  Fig. 5 presents the hit point difference against UtalFighter by our variant bots. The curve is averaged by the latest 50 round results at each epoch and it tends to get converge over time. Larger hp difference indicates the bot is more competitive  to a certain degree. Though RHEAOM-SL improves itself steadily, RHEAOM-Q and RHEAOM-PG both have achieved better performance at the early phase. Besides, RHEAOM-PG shows better than RHEAOM-Q. The trend of RHEAOM-R is other than other variants of RHEAOM and its curve is still varying randomly, but RHEAOM-R also has a larger hp difference than RHEA. Because of the absence of the opponent model, RHEA has the worst performance among all variants.

\begin{figure}[H]
	\centering
	\includegraphics[width=8cm, height=4.5cm]{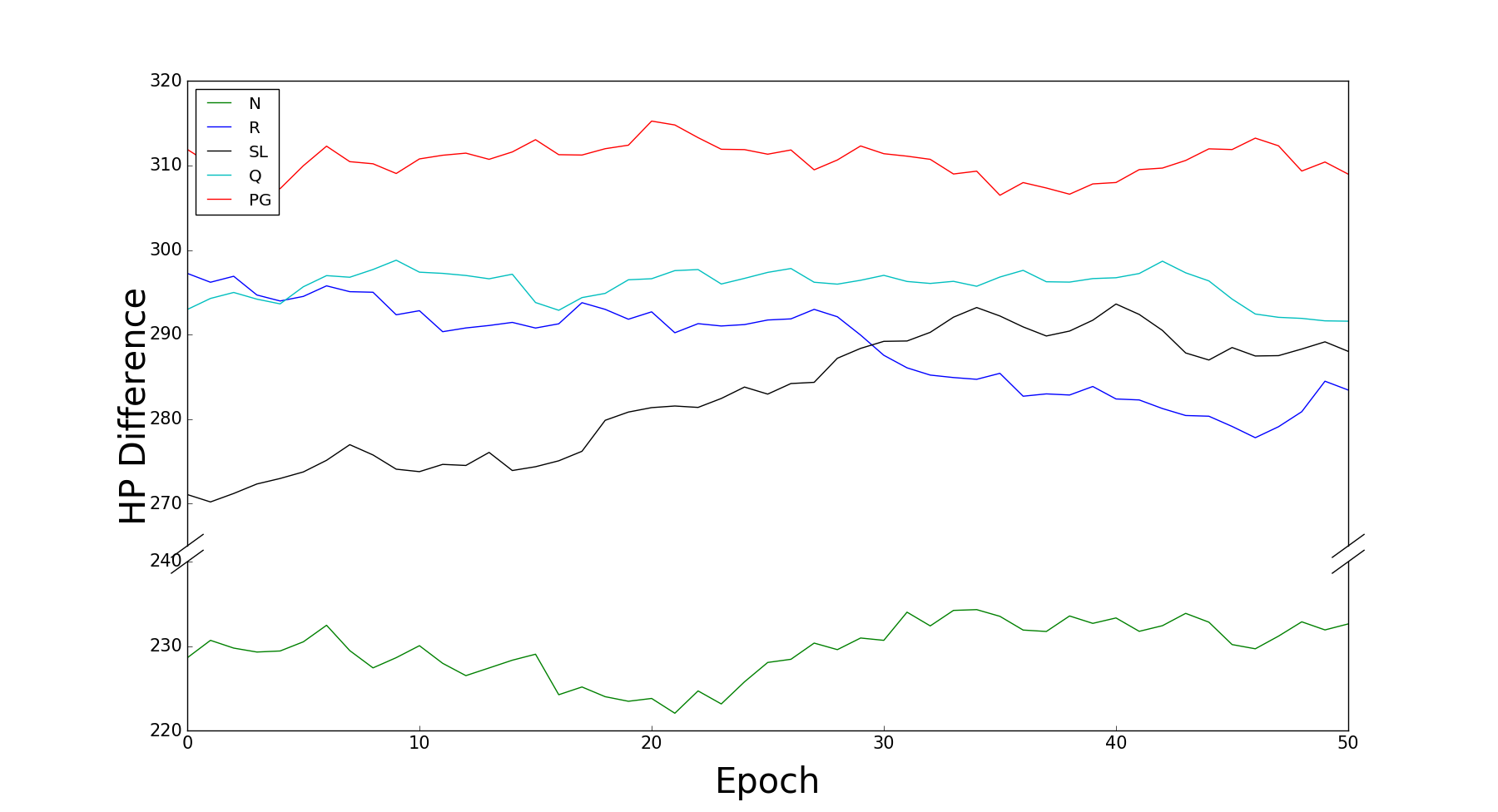}
	\caption{The curves of the mean hp difference means RHEA-based bots fight against UtalFighter with the increasing of iterations.}
	\label{pic:plot_hp}
\end{figure}

According to Table \ref{tab:vs-2018}, RHEA is able to win all bots from 2018 FTGAIC for all three characters. However, RHEAOM-R performs worse than RHEA especially in character ZEN, since the random opponent model may ruin the evaluation process of the rolling horizon and mislead the bot to inappropriate decisions. The combination with supervised-learning-based and reinforcement-learning-based opponent model both result in significant improvement to RHEA. For instance, RHEA is only slightly better than UtalFighter while all RHEAOMs defeat UtalFighter in more than 85 percent of games.

All variants of RHEAOM show competitive performance against all opponents. 
Opponent model treats opponent as part of the observation and can respond to adaptive opponents. And the change in policy is explicitly encoded by the neural network model. RHEAOM-PG achieves the best performance since it not just mimic the opponent’s behavior but also find out the most advantageous action from the opponent. 

Since all these strong opponent agents are based on MCTS and the time is limited in a real-time video game, it cannot guarantee that the optimization process has converged to an optimal solution as sampling-based optimization methods require many simulations. 
Compared with MCTS, RHEA is more efficient because of its simplicity and the correlation in action sequence. In addition, the fighting game is a dynamic process without a fixed optimal strategy to win every game. The bot should give a dynamic adaptive response to opponent's behavior, thus it cannot ensure to obtain a suitable response towards opponent unless one side is always defeated.

\subsection{Comparisons between RHEAOM and MCTSOM}

In order to inspect the adaption of our opponent learning model for other statistical forward planning algorithms such as Monte-Carlo Tree Search, we set up the comparison experiment between RHEAOM and MCTSOM. From the above experiments, supervised-learning-based opponent model and policy-gradient-based opponent model show the best performance among RHEA variants, so these opponent models are introduced into Thunder, which is the strongest MCTS-based fighting bot in above opponents. We call the variants of Thunder are ThunderOM-SL (supervised-learning-based) and ThunderOM-PG (policy-gradient-based) respectively.

The results of RHEAOM against ThunderOM are presented in Table \ref{tab:mctsom}. In terms of  winning rate, ThunderOMs have been improved in all three characters when compared with Thunder, and ThunderOM-PG is slightly better than RHEAOMs when the character is ZEN. The results indicate that our proposed opponent learning model is well suitable for the statistical forward planning algorithms, not only RHEA but also MCTS.           

\begin{table}[h]
	\centering
	\begin{tabular}{c|C{0.65cm}C{0.65cm}C{0.65cm}|C{0.65cm}C{0.65cm}C{0.65cm}}
		\toprule
		\diagbox{Opps}{Ours}  & \multicolumn{3}{c|}{RHEAOM-SL(\%)} & \multicolumn{3}{c}{RHEAOM-PG(\%)} \\
		\midrule
		Chars   & GAR   & LUD   & ZEN   & GAR   & LUD   & ZEN \\
		\midrule
		\multicolumn{1}{c|}{ThunderOM-SL} & 75.3(4)  & 68.4(5)  & \multicolumn{1}{c|}{58.1(6) } & 81.1(4)  & 69.6(2)  & 54.2(4) \\
		\multicolumn{1}{c|}{ThunderOM-PG} & 77.2(3)  & 55.6(4)  & \multicolumn{1}{c|}{49.6(4) } & 79.1(5)  & 60.3(3)  & 48.3(3)  \\
		\midrule
		Thunder  & 84.0(2)  & 83.6(2)  & 69.8(2)  & 89.8(4)  & 89.3(1)  & 76.4(3)  \\
		\bottomrule
	\end{tabular}%
	\caption{\textbf{Win rate against ThunderOMs.} Every test repeated for five times to average the win rate against MCTS with supervised-learning-based and policy-gradient-based opponent model. The \texttt{Mean} row denotes the average win rates against ThunderOM bots. The values in parentheses denote the 95\% confidence interval.}
	\label{tab:mctsom}%
\end{table}%

 \subsection{Results on 2019 FTGAIC}

To further verify the performance of our proposed framework for fighting game AI, according to the above experimental results, we choose Rolling Horizon Evolution with Policy-gradient based opponent learning model (named RHEAPI, but actually it is RHEAOM-PG) to participate the 2019 FTGAIC, which is sponsored by 2019 IEEE Conference on Games (CoG). 

\begin{table}[htbp]
	\centering
	\begin{tabular}{ccc}
		\toprule
		\multicolumn{1}{c}{Name of Bot} & \multicolumn{1}{c}{Score} & \multicolumn{1}{c}{Rank} \\
		\midrule
		ReiwaThunder & 133   & 1 \\
		\textbf{RHEAPI (ours)} & 122   & 2 \\
		Toothless & 91    & 3 \\
		FalzAI & 68    & 4 \\
		LGISTBot & 67    & 5 \\
		SampleMctsAi (baseline) &	52	&6 \\
		HaibuAI &	32	&7 \\
		DiceAI &	19	&8 \\
		MuryFajarAI	&17	&9 \\
		TOVOR	&9	&10 \\
		
		\bottomrule
	\end{tabular}%
	\caption{\textbf{Rank of 2019 FTGAIC.}}
	\label{tab:2019-FTG-rank}%
	
\end{table}%

\begin{figure*}[htbp]
	\centering
	\begin{subfigure}{0.32\linewidth}
		\includegraphics[width=6cm, height=3cm]{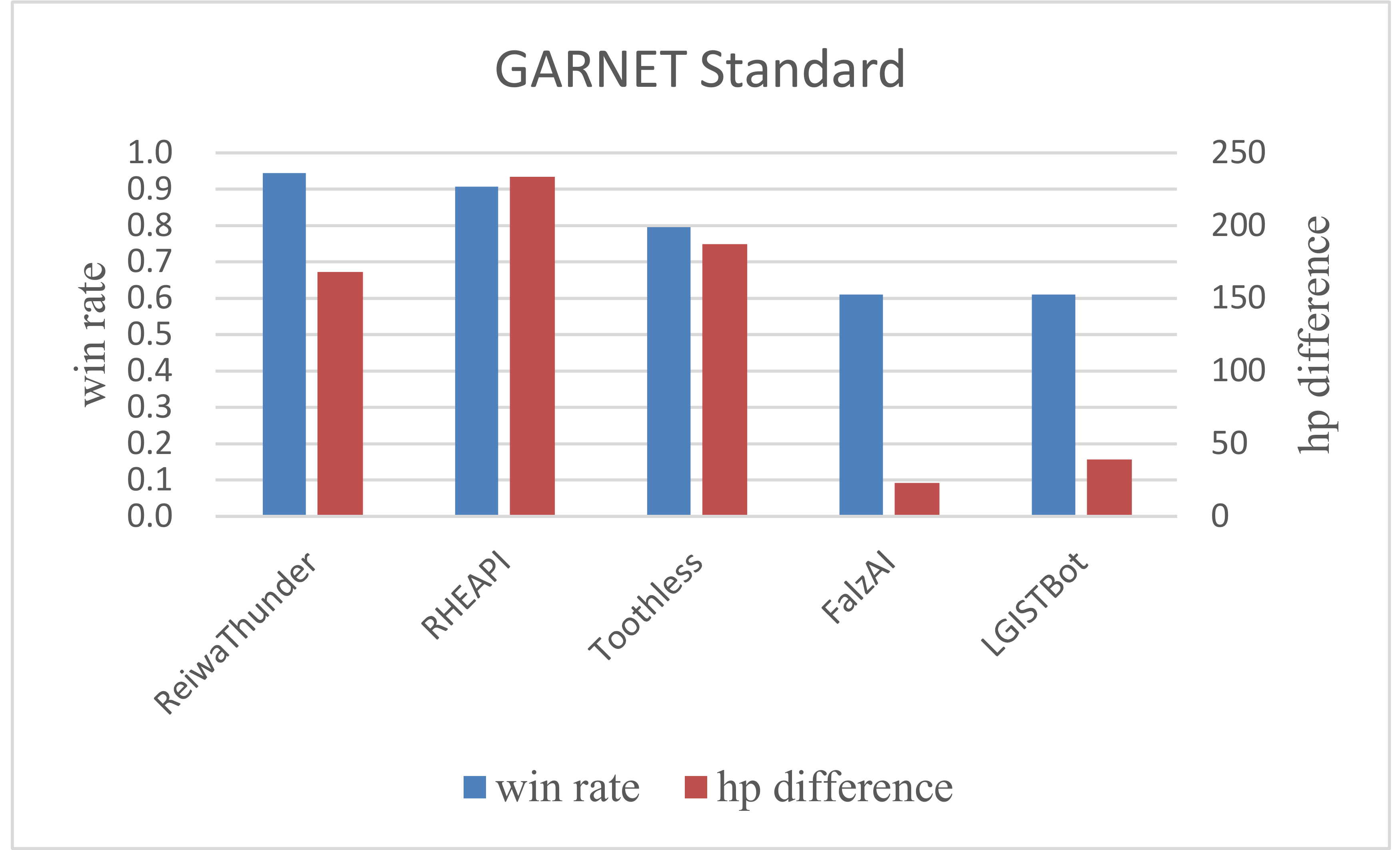}
	\end{subfigure}
	\begin{subfigure}{0.32\linewidth}
		\includegraphics[width=6cm, height=3cm]{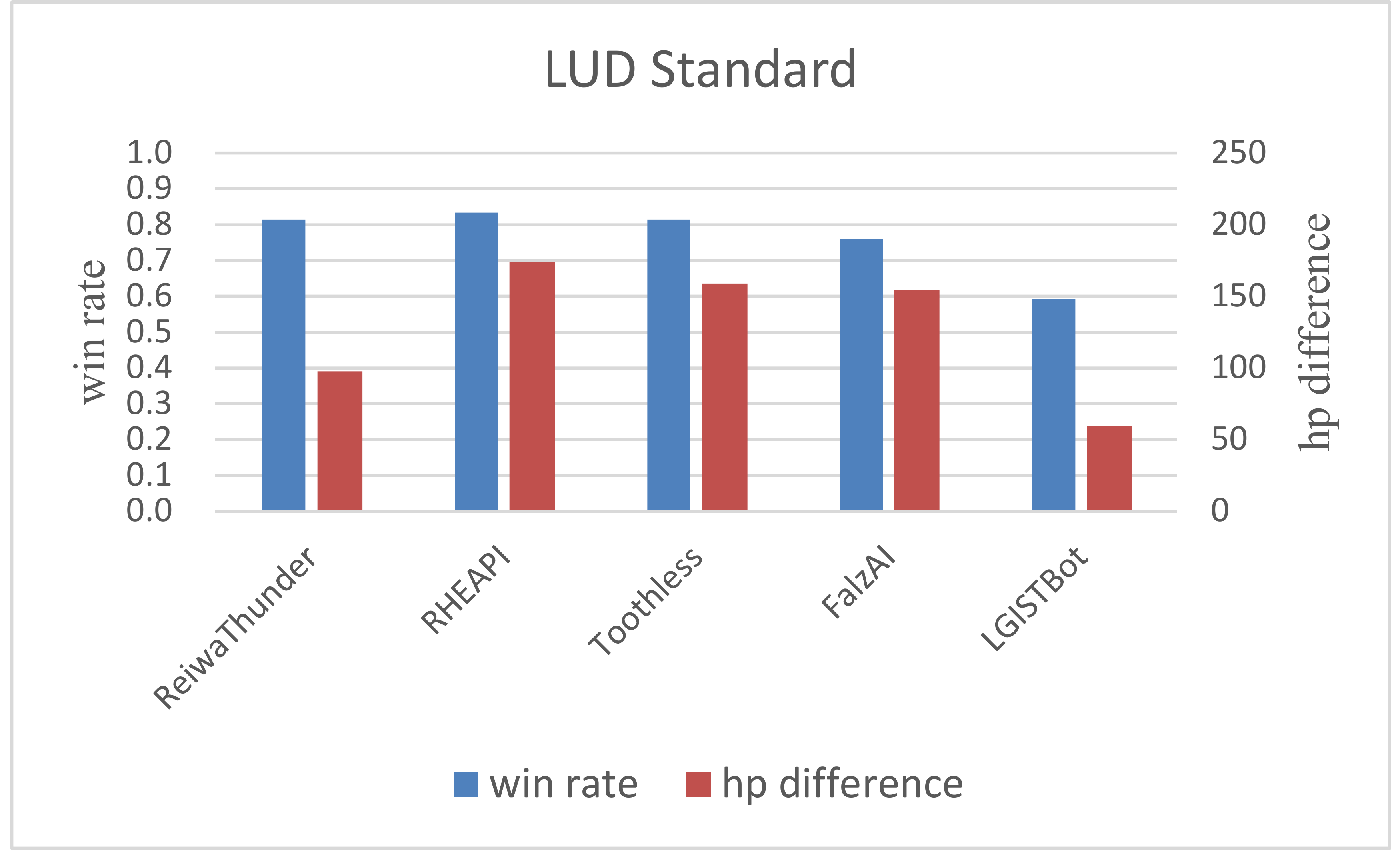}
	\end{subfigure}
	\begin{subfigure}{0.32\linewidth}
		\includegraphics[width=6cm, height=3cm]{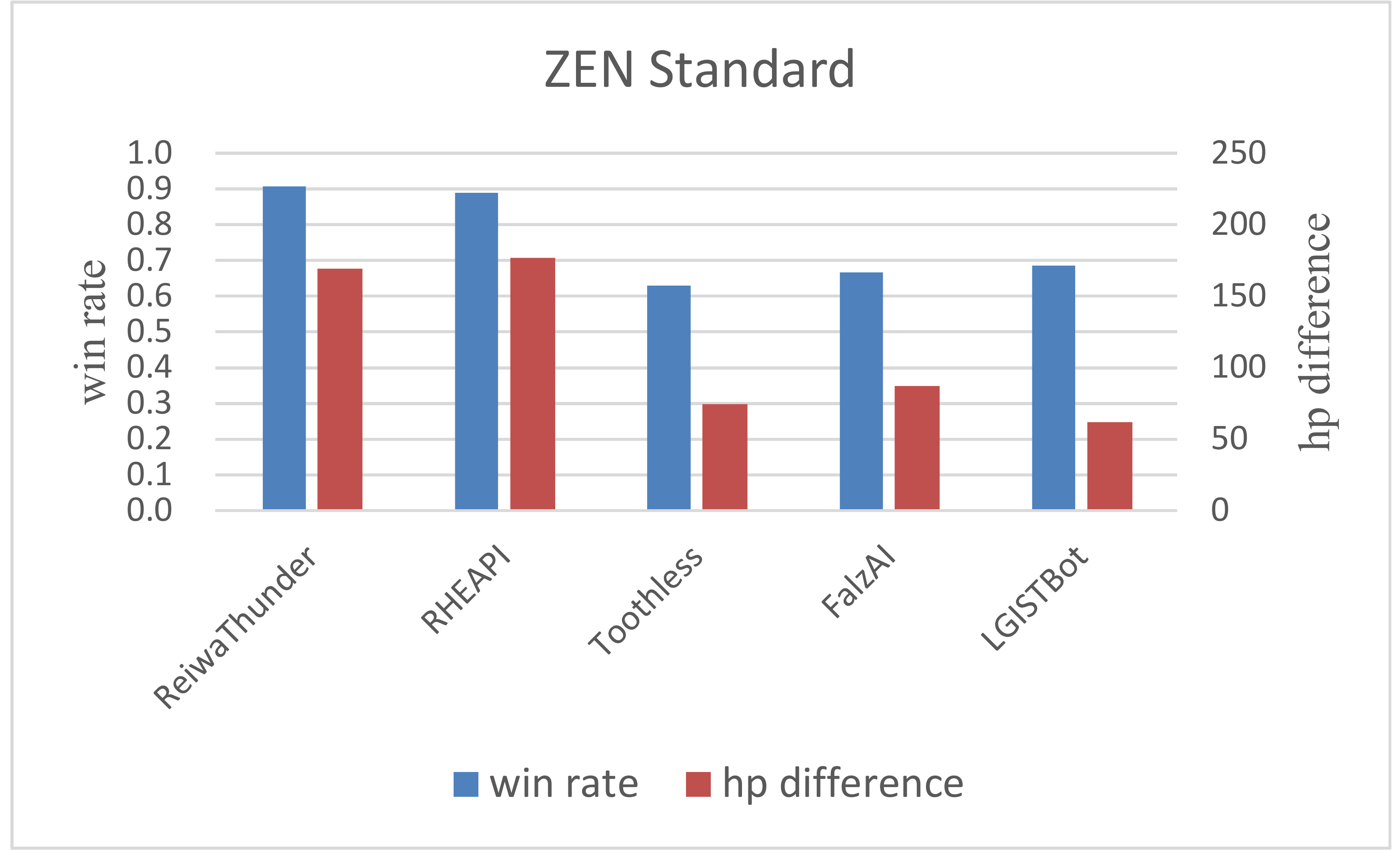}
	\end{subfigure}
	
	\begin{subfigure}{0.32\linewidth}
		\includegraphics[width=6cm, height=3cm]{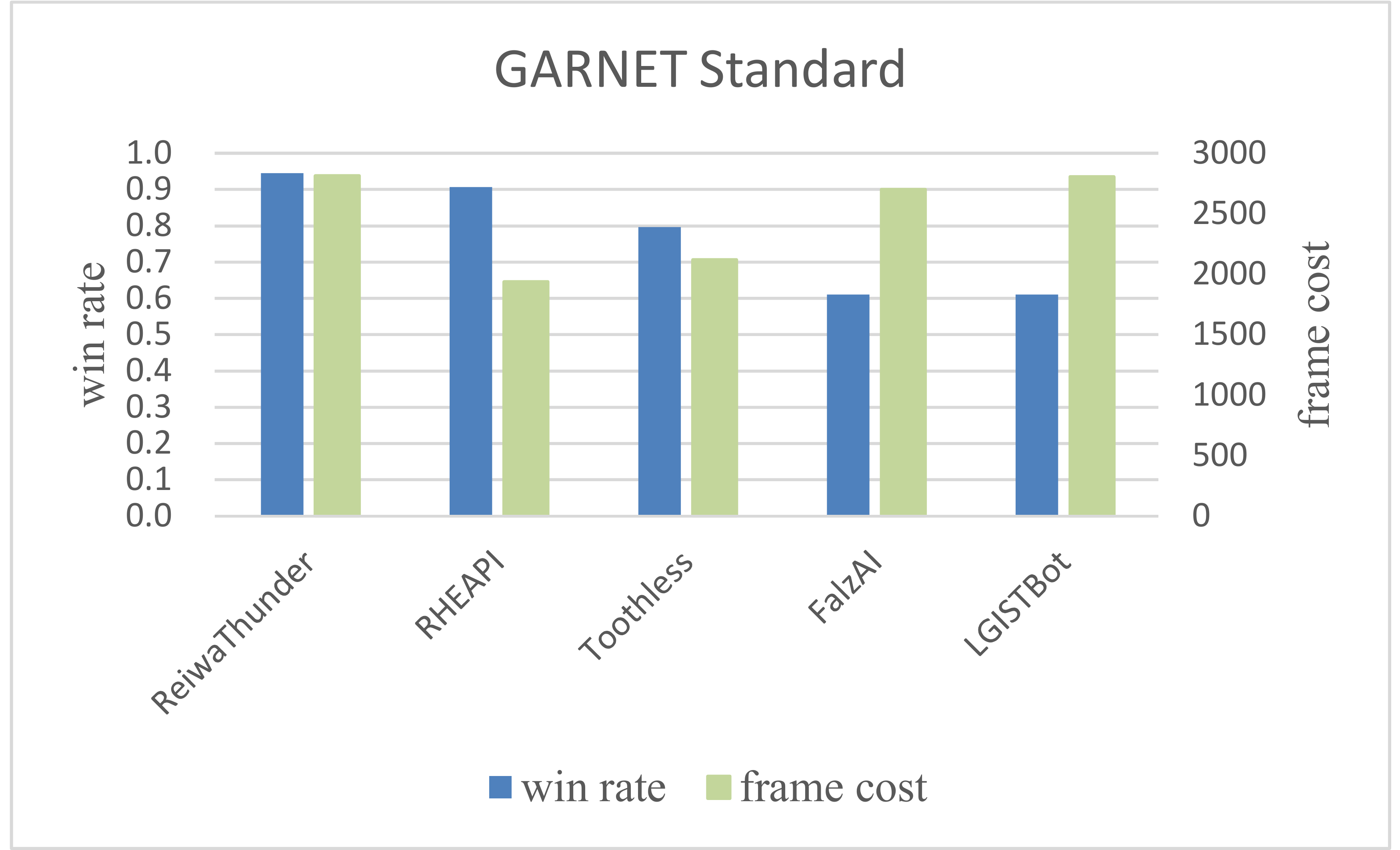}
	\end{subfigure}
	\begin{subfigure}{0.32\linewidth}
		\includegraphics[width=6cm, height=3cm]{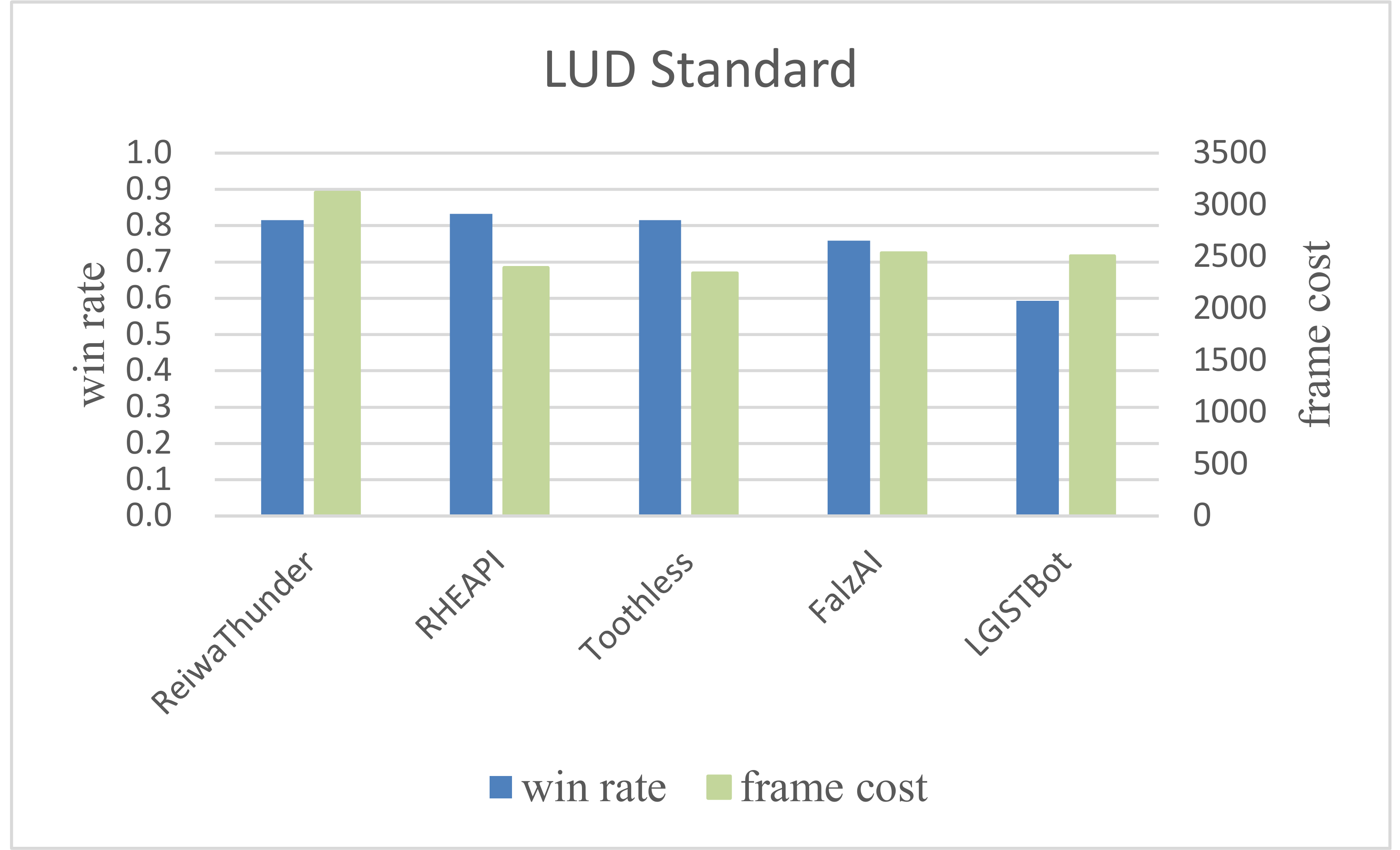}
	\end{subfigure}
	\begin{subfigure}{0.32\linewidth}
		\includegraphics[width=6cm, height=3cm]{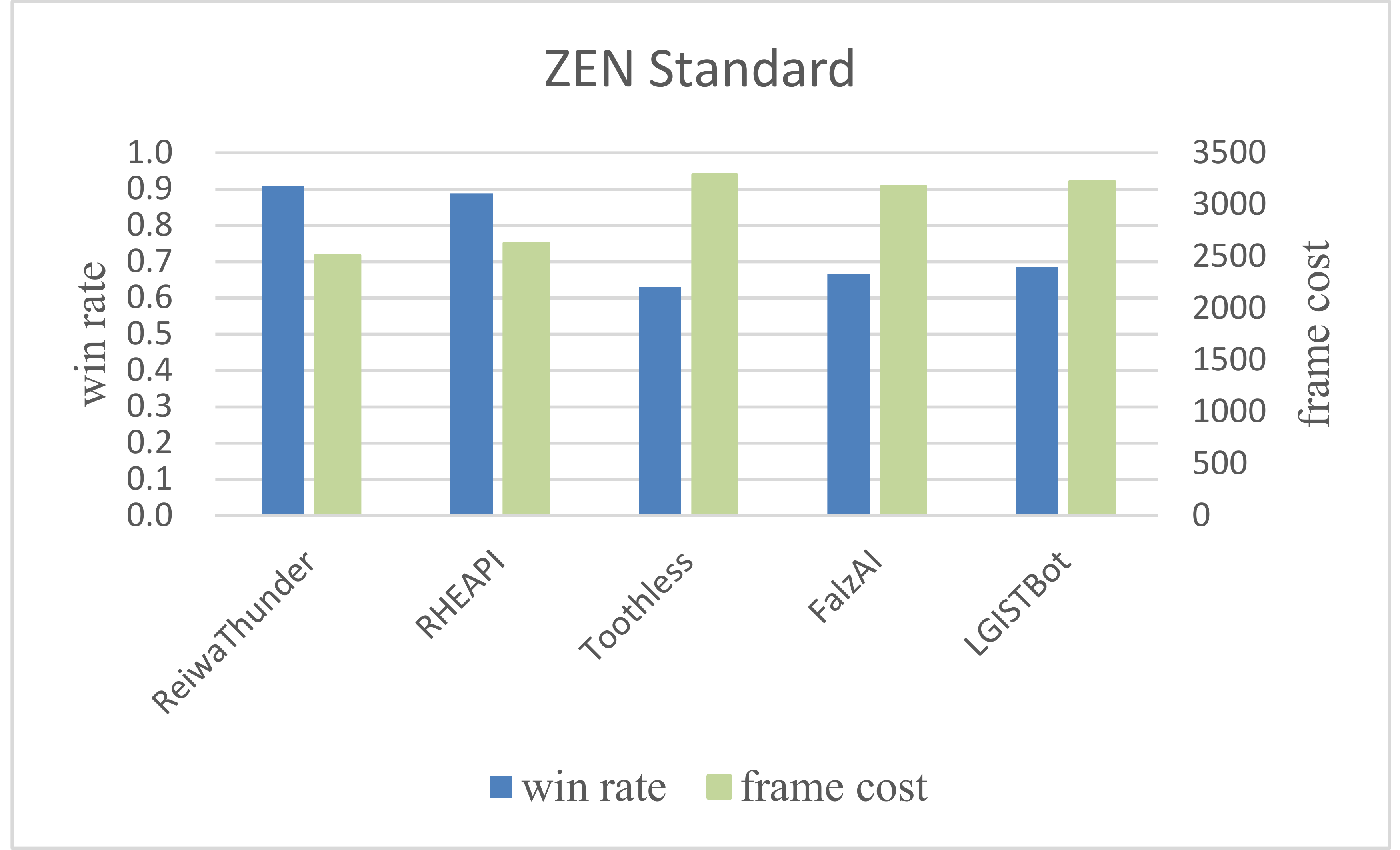}
	\end{subfigure}
	\caption{Win rate compared with hp difference and frame cost of 2019 FTGAIC top five bots in Standard League.} \label{2019-standard}
	
	\begin{subfigure}{0.32\linewidth}
		\includegraphics[width=6cm, height=3cm]{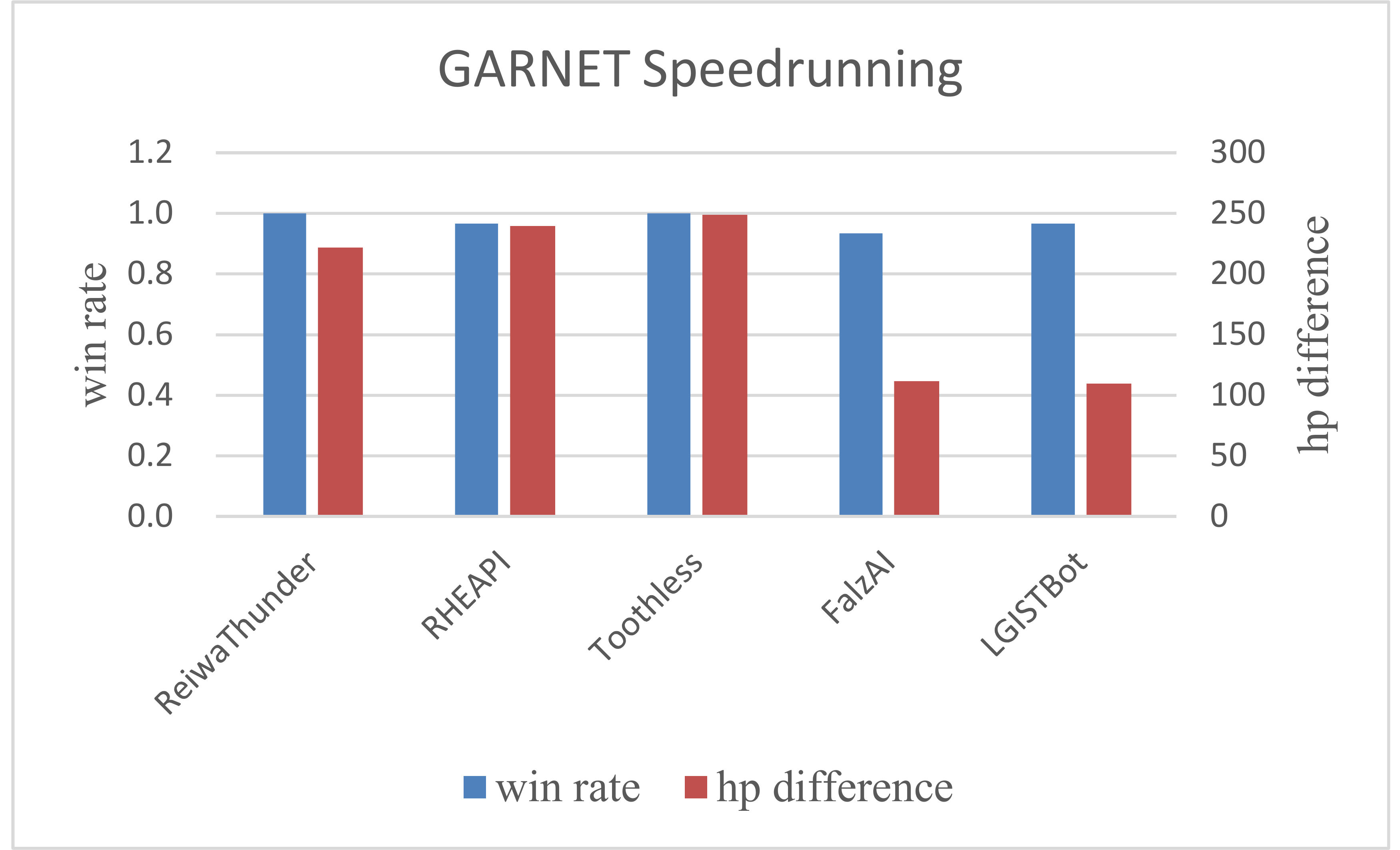}
	\end{subfigure}
	\begin{subfigure}{0.32\linewidth}
		\includegraphics[width=6cm, height=3cm]{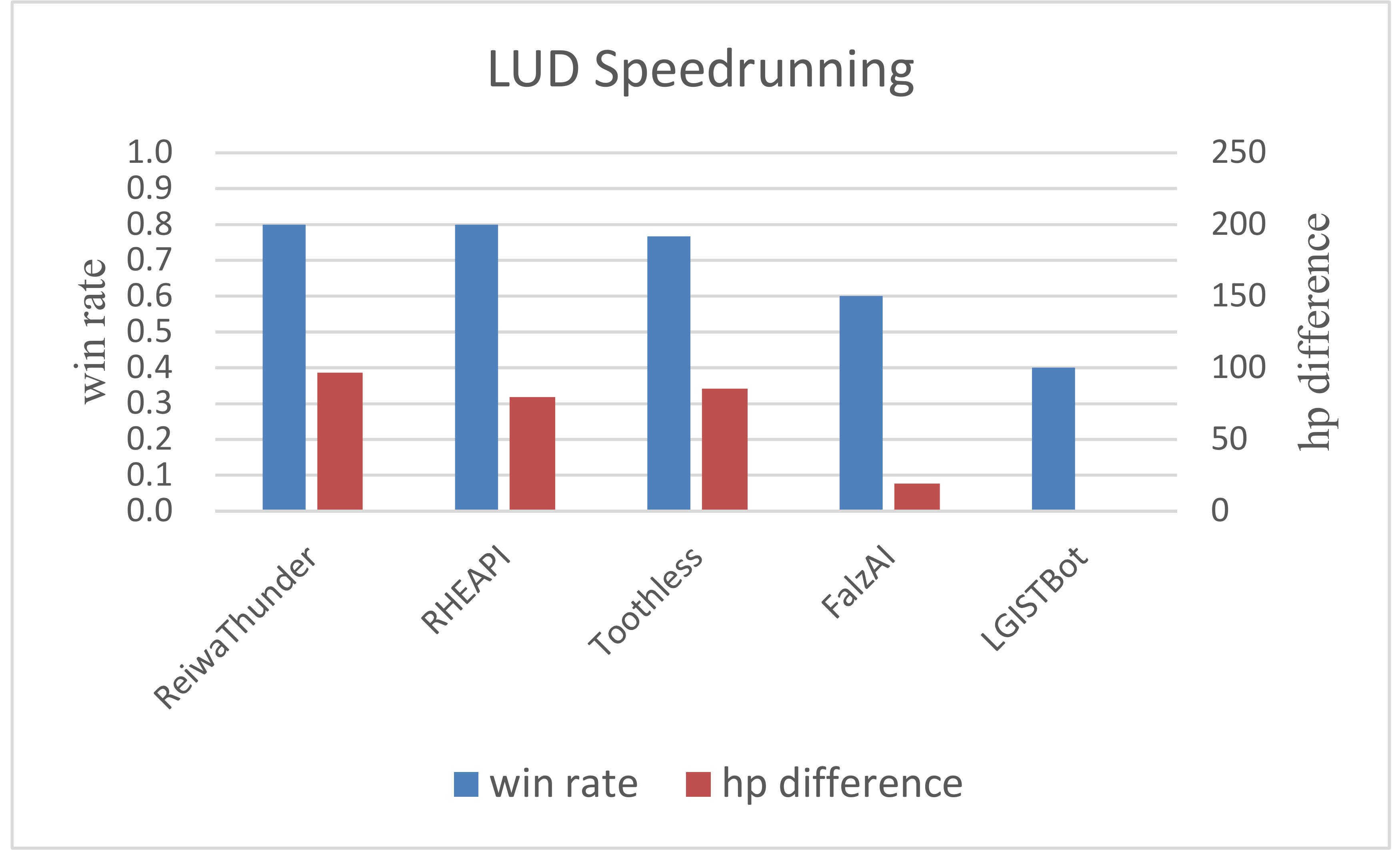}
	\end{subfigure}
	\begin{subfigure}{0.32\linewidth}
		\includegraphics[width=6cm, height=3cm]{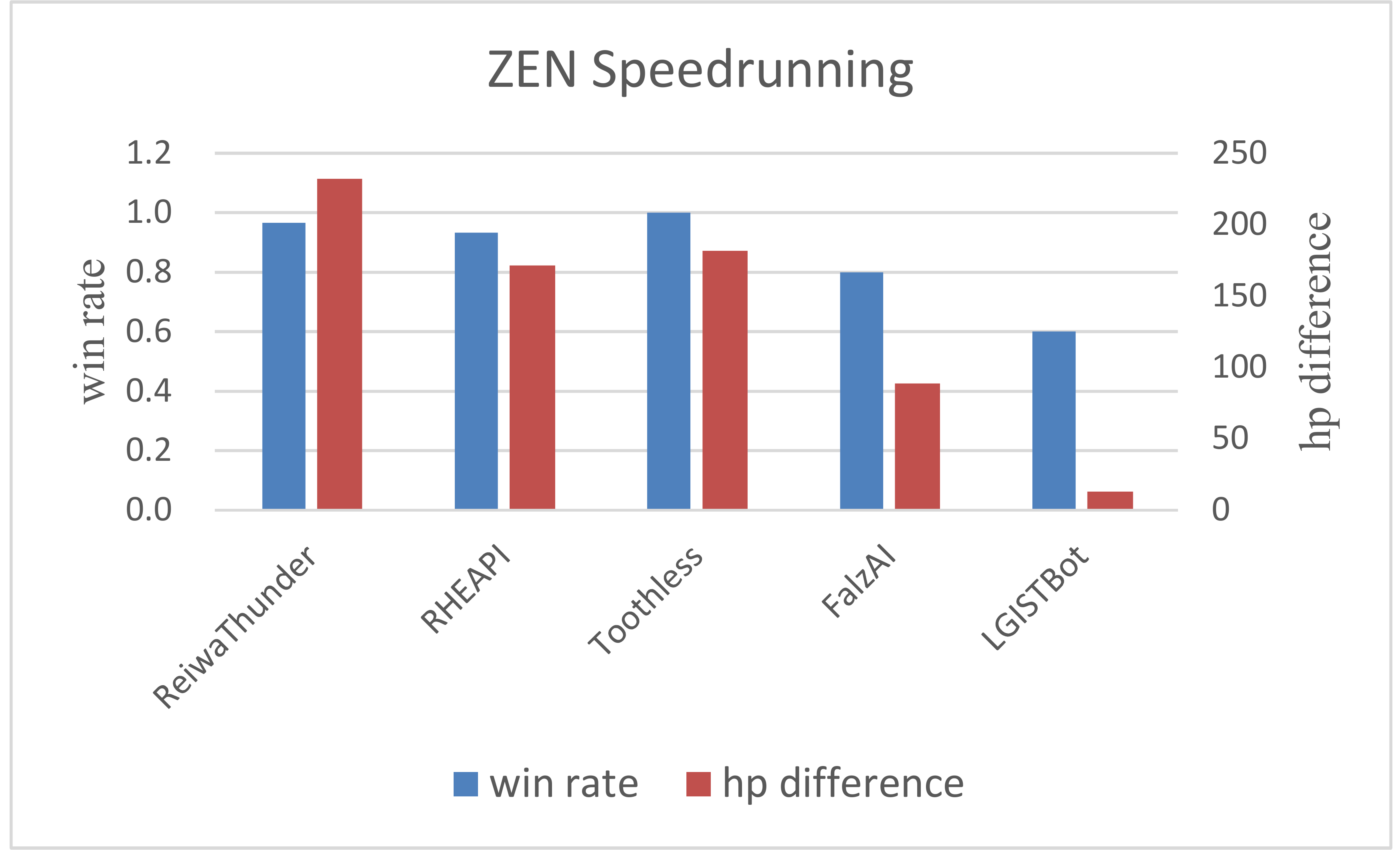}
	\end{subfigure}
	
	\begin{subfigure}{0.32\linewidth}
		\includegraphics[width=6cm, height=3cm]{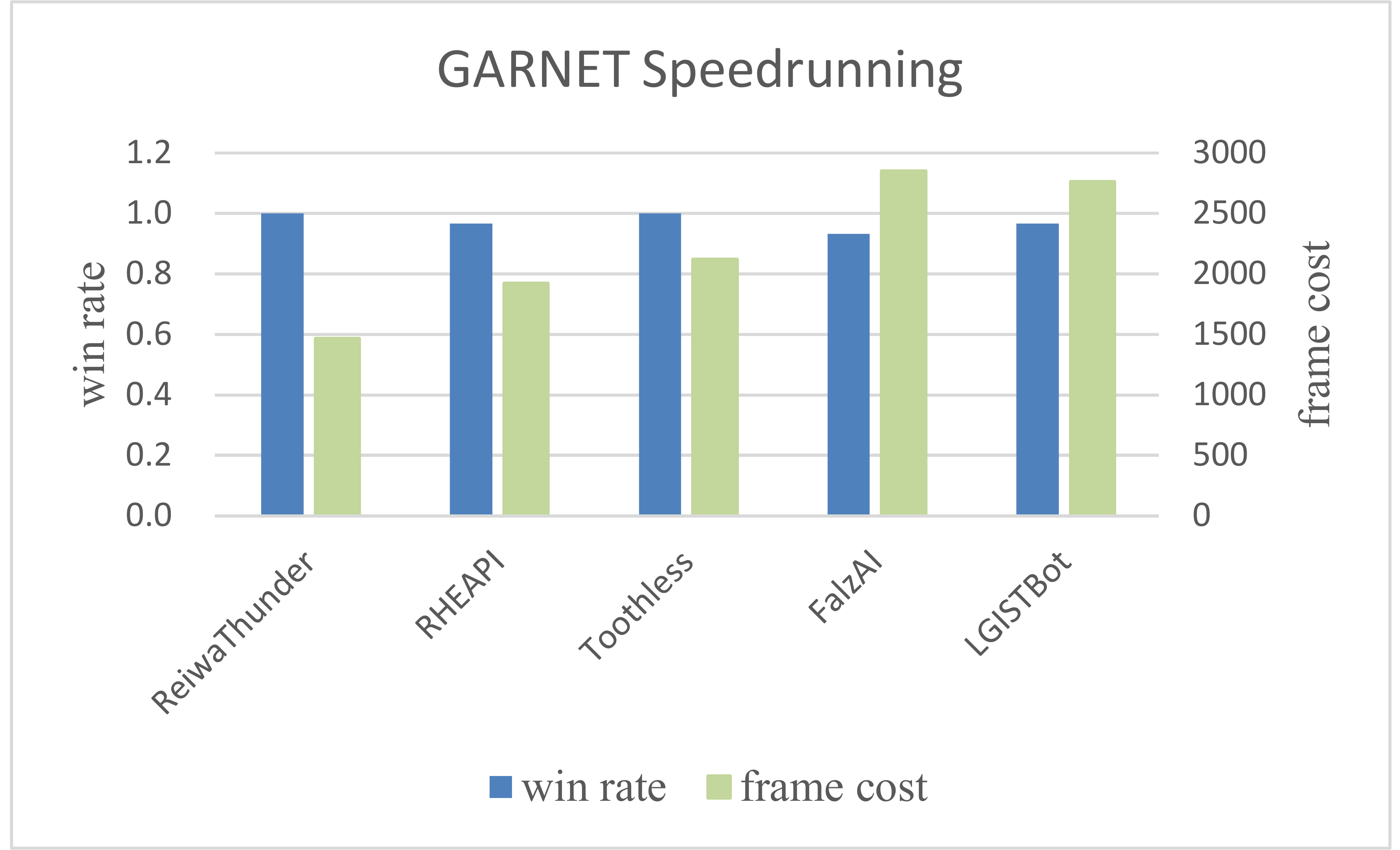}
	\end{subfigure}
	\begin{subfigure}{0.32\linewidth}
		\includegraphics[width=6cm, height=3cm]{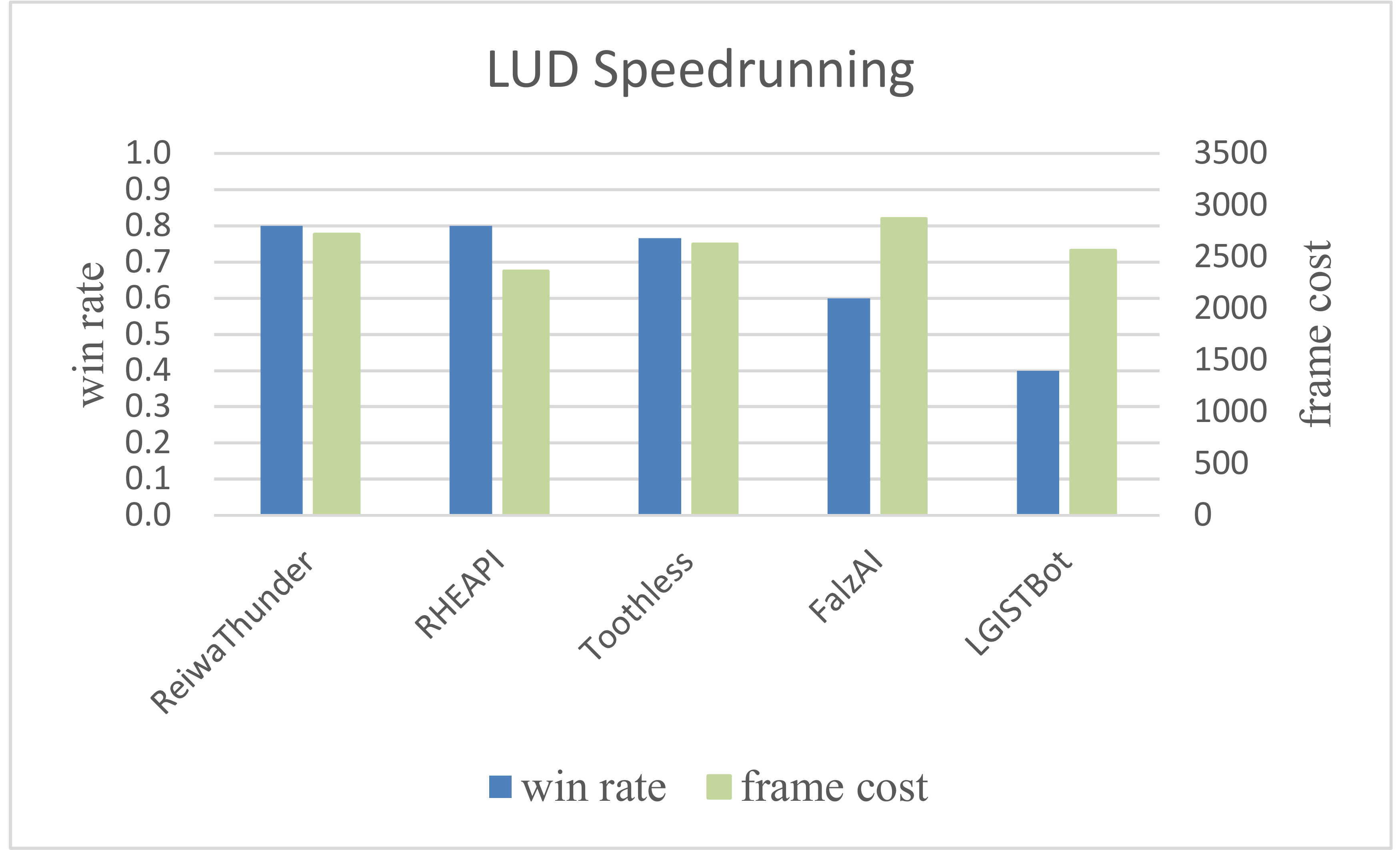}
	\end{subfigure}
	\begin{subfigure}{0.32\linewidth}
		\includegraphics[width=6cm, height=3cm]{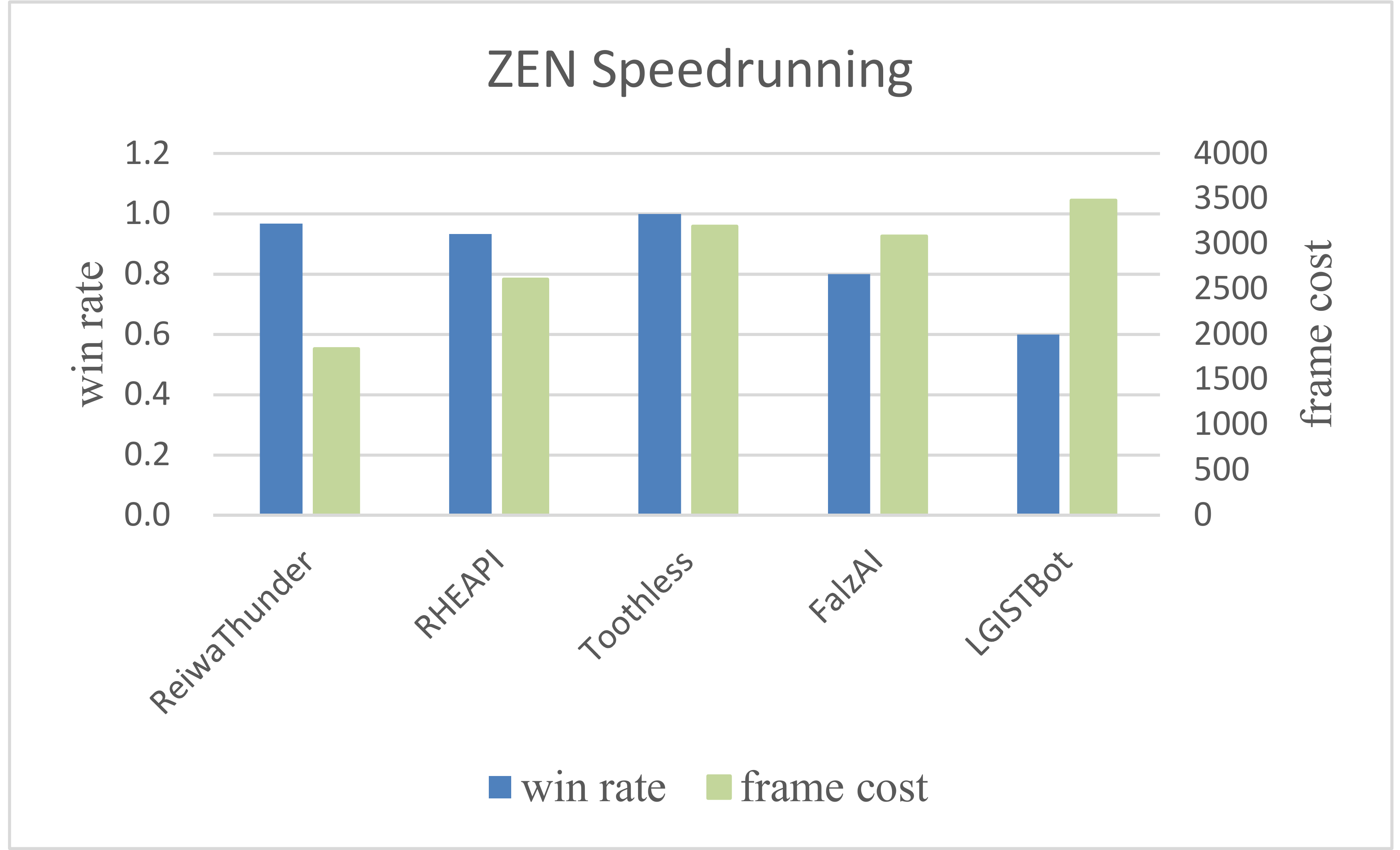}
	\end{subfigure}
	
	\caption{Win rate compared with hp difference and frame cost of 2019 FTGAIC top five bots in Speedrunning League.}\label{2019-speed}
\end{figure*}

\textbf{Performance in competition.} There are a total of 10 bots participating in this competition. As presented in Table \ref{tab:2019-FTG-rank}, ReiwaThunder wins the first place, which is an improved version of 2018 champion Thunder. While our RHEAPI wins the runner-up of the competition and its score comes very close to the first place. According to official statistics, RHEAPI wins two or three games less than ReiwaThunder and wins over the ReiwaThunder for character LUD, whose action data is not available in advance. It demonstrates that RHEA with an opponent learning model is a competitive unified framework for fighting game AI. Here we list the brief description of top five bots in 2019 FTGAIC:
\begin{itemize}
	\item \textbf{\emph{ReiwaThunder}}, $1^{st}$, based on Thunder while replacing MCTS with MiniMax and a set of heuristic rules for each character. 
	\item \textbf{\emph{RHEAPI}}, $2^{nd}$, RHEA combines with policy-gradient-based opponent model.
	\item \textbf{\emph{Toothless}} \cite{toothless}, $3^{rd}$, based on KotlinTestAgent and with a combination of MiniMax, MCTS and some basic rules. 
	\item \textbf{\emph{FalzAI}}, $4^{th}$, MCTS-based bot combines with switchable general strategy, including aggressive and defensive.
	\item \textbf{\emph{LGISTBot}} \cite{lgistbot}, $5^{th}$, hybrid methods with MCTS and genetic action sequence.
\end{itemize}

Note that there are two leagues to fully verify the performance of participants. The Standard League is set for competition among all bots, and the winner is the one with the maximum of average win rate against all other bots. The Speedrunning League is set for fighting against official bot SampleMctsAi, and the winner is to beat the SampleMctsAi with the shortest average time. In order to directly evaluate our agent, we focus on the three key factors win rate, hp difference, and frame cost from the competition logs. The competition results of the top five bots in 2019 FTGAIC are shown in Fig. \ref{2019-standard} and Fig. \ref{2019-speed}.

The win rate of RHEAPI is quite close to ReiwaThunder whether in the Standard League or the Speedrunning League. In the Standard League, RHEAPI causes the largest amount of hp difference to the opponents than other bots.  Beyond that, the average frame cost of RHEAPI is the lowest among all bots for three characters. It indicates that RHEAPI is more aggressive and effective to beat the opponents. When in the Speedrunning League, RHEAPI spends more frames to beat the baseline bot than ReiwaThunder except for in character LUD. It means that heuristic knowledge of ReiwaThunder still plays an important role when fights against the specific and known opponent. On the whole, there is a  negative correlation between hp difference and frame cost to a certain extent. 


\subsection{Discussion}
RHEA is an efficient statistical forward planning approach, and the goal of RHEA is to search for the best action sequence for decision and planning. However, fighting game is a real-time two-player zero-sum game, so it is insufficient to consider only one-side action sequence and neglect the other side. 

To address this deficiency, we propose three variants of opponent learning models RHEAOM-SL, RHEAOM-Q, and RHEAOM-PG respectively. RHEAOM-SL directly mimics the opponent's behaviors in the game.  However, it is easily to be misled when the opponent adopts unsuitable response to the other side.    
Unlike supervised learning, reinforcement learning does not directly learn the mapping rule of state-action pairs but instead learns the optimal opponent policy with reward signals. From this point of view, we propose two opponent learning models that are based on two effective reinforcement learning approaches: Q-learning and policy gradient. According to the Fig. \ref{pic:plot}, compared to the overestimation or underestimation problems in Q-learning, policy gradient-based approach finds the optimal opponent policy more rapidly and steadily, which leads strong performance of RHEAOM-PG in the fighting game.  

Constrained by the limit energy, it suggests the uneven distribution of the generated state-action pairs from fighting game. Since each character has a limited amount of energy, and hence using this wisely is a challenge for intelligent game play. Reinforcement learning balances accurate opponent modelling versus creating an opponent that plays well. 
For instance, the occurrence of deadly skills is far less than those common actions without an energy cost. Since the supervised-learning-based opponent model is able to obtain a higher prediction accuracy than reinforcement-learning-based opponent models, it mainly infers common actions but not the deadly skill. However, owing to the reward is a measuring signal, it can represent the varying level of importance for actions, and make the inference of reinforcement-learning-based opponent model more effective than the supervised-learning-based.



\section{Conclusion \& Future Work} \label{conclusion}
This paper presents RHEAOM, a novel fighting AI framework that utilizes evolutionary strategy and opponent modeling to search the best action sequence for a real-time fighting game. In our work, we propose three variants of RHEAOM, which are RHEAOM-SL, RHEAOM-Q, RHEAOM-PG. Within the aid of opponent model,  RHEAOM is able to outperform the state-of-the-art MCTS-based fighting bots. Experiment results suggest that our method can efficiently find the weakness of opponents and select competitive actions for all three characters in the 2018 FTGAIC. Moreover, RHEAOM-PG becomes the runner-up of FTGAIC in 2019 IEEE CoG.

Even though RHEAOM has achieved impressive performance in experiments, it still cannot completely defeat all opponent in the competition. We will consider to introduce a  model-based deep reinforcement learning method, instead of using the built-in forward model, to improve the adaption and generalization of the whole learning algorithm. 
Besides, it is not easy for an opponent model to accurately predict the actions of the opponent because it mainly depends on the predictability of the opponent and is restricted by the real-time constraint. We will give more investigation on this research topic.

Although the results in this paper are all on thr FightingICE platform, the approach of enhancing RHEA with a learned opponent model is generally applicable to any two player game.   The only part of the system that is specific to FightingICE is the set of 18 features used as input to the opponent model neural network.   We plan to also test the RHEAOM methods on other two-player real-time video games, such as Planet Wars \cite{LucasPlanetWars}.

An interesting and important challenge is to make the approach more general, while still achieving the same degree of very rapid learning.  It is worth emphasizing that our method learns the opponent model from scratch after the first round of play, and all this is conducted within the constraints of a real-time tournament.

\ifCLASSOPTIONcaptionsoff
  \newpage
\fi



\bibliographystyle{IEEEtran}
\bibliography{ref}
\end{document}